Thesis for the Degree of Masters of Science

# Prediction Model of Aqua Fisheries Using IoT Devices

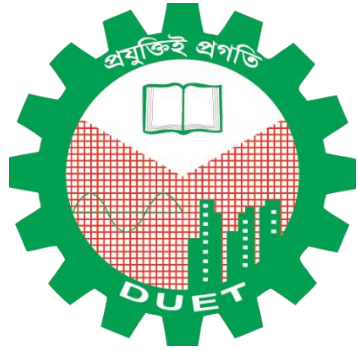

**Md. Monirul Islam**
Student ID: 16204025-P

**Department of Computer Science and Engineering**

**Dhaka University of Engineering & Technology (DUET), Gazipur**

**Gazipur-1707, Bangladesh**

**June 2021**

# Prediction Model of Aqua Fisheries Using IoT Devices

by

**Md. Monirul Islam**

Student ID: 16204025-P

Supervised by

**Prof. Dr. Mohammod Abul Kashem**

The thesis submitted to the Department of Computer Science and Engineering of Dhaka University of Engineering & Technology, Gazipur in partial fulfillment of the requirements for the Degree of Masters of Science in Computer Science and Engineering

The Thesis entitled "Prediction Model of Aqua Fisheries Using IoT Devices" submitted by Md. Monirul Islam, Student No.:16204025-P has been accepted as satisfactory in partial fulfillment of the requirement for the Degree of Master of Science in Computer Science and Engineering on June 15, 2021.

## BOARD OF EXAMINERS

1. _______________________
   (Dr. Mohammod Abul Kashem)   (Supervisor)
   Professor                     (Chairman)
   Department of Computer Science and Engineering
   Dhaka University of Engineering & Technology, Gazipur
   Gazipur-1707, Bangladesh.

2. _______________________
   (Dr. Md. Fazlul Hasan Siddiqui)   Member
   Professor                          Ex-officio
   Department of Computer Science and Engineering
   Dhaka University of Engineering & Technology, Gazipur
   Gazipur-1707, Bangladesh

3. _______________________
   (Dr. Md. Nasim Akhtar)   Member
   Vice Chancellor
   Chandpur Science and Technology University &
   Professor, Department of Computer Science and Engineering
   Dhaka University of Engineering & Technology, Gazipur
   Gazipur-1707, Bangladesh

4. _______________________
   (Dr. Momotaz Begum)   Member
   Associate Professor
   Department of Computer Science and Engineering
   Dhaka University of Engineering & Technology, Gazipur
   Gazipur-1707, Bangladesh.

5. _______________________
   Dr. Mohammad Kaykobad   Member
   Former Professor         (External)
   Department of Computer Science & Engineering
   Bangladesh University of Engineering & Technology (BUET),
   Dhaka, Bangladesh.

# DECLARATION

I hereby declare that the work presented in this thesis or any part of this thesis has not been submitted elsewhere for the award of any degree or diploma. Information derived from the published and unpublished work of others has been acknowledged in the text and a list of references is given.

Signature

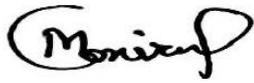

(Md. Monirul Islam)
Student No.16204025-P

Signature of Thesis Supervisor

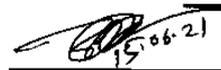

**(Dr. Mohammod Abul Kashem)**
Professor
Department of Computer Science and Engineering
Dhaka University of Engineering & Technology, Gazipur
Gazipur-1707, Bangladesh

# DEDICATION

*To my parents,*

*Mr. Md. Najar Sheikh and Sanuara Begum*

# ACKNOWLEDGMENT

All praises are for the Almighty Allah for giving me the strength, without which I could not afford to attempt this research work. I would like to express my gratitude to my honorable supervisor, Professor Dr. Mohammod Abul Kashem, Department of Computer Science and Engineering, Dhaka University of Engineering and Technology (DUET), Gazipur, for his relentless guidelines, knowledge, and expertise throughout my whole study period. It has been my honor to work with the supervisor that helps me to increase my research knowledge. I would like to thank all the members of the board of examiners for their precious time in reviewing my work and their valuable comments. It is a pleasure to thank those who made this thesis possible. The thesis work was completed at the Department of Computer Science and Engineering, Dhaka University of Engineering and Technology (DUET), Gazipur. In addition, I would like to express my thanks to Dr. Jia Uddin, Assistant Professor, Technology Studies Department, Endicott College, Woosong University, Daejeon, South Korea for his cooperation. I am also grateful to my parents, without their trust and support, I would have never arrived here.

<div style="text-align: right;">
Md. Monirul Islam<br>
June 2021
</div>

# ABSTRACT

Aquaculture is the cultivation of natural, regulated marine and freshwater aquatic creatures. The real-time monitoring of aquatic environmental parameters is very important in fish farming. Internet of Things (IoT) can play a vital role in real-time monitoring. This thesis presents an IoT framework for the efficient monitoring and effective control of different aquatic environmental parameters related to the water. The proposed system is implemented as an embedded system using sensors and an Arduino. Different sensors including pH, temperature, and turbidity are placed in cultivating pond water and each of them is connected to a common microcontroller board built on an Arduino Uno. The sensors read the data from the water and store it as a CSV file in an IoT cloud named Thingspeak through the Arduino Micro-controller. In the experimental part, we have collected data from 5 ponds with various sizes and environments. After getting the real-time data, we have compared these with the standard reference values. As a result, we can take the decision which ponds are satisfied for cultivating fish and what is not. After that, we have labeled the data with 11 fish categories including katla, sing, prawn, rui, koi, pangas, tilapia, silvercarp, karpio, magur, and shrimp. In addition, the data were analyzed using 10 machine learning (ML) algorithms containing J48, Random Forest, K-NN, K*, LMT, REPTree, JRIP, PART, Decision Table, and Logit boost. After experimental evaluation, it was observed among 5 ponds, only three ponds were perfect for fish farming, where these 3 ponds only satisfied the standard reference values of pH (6.5-8.5), Temperature (16-24)°C, Turbidity (below 10)ntu, Conductivity (970-1825)μS/cm, and Depth (1-4) meter. Among the state-of-art machine learning algorithms, Random Forest achieved the highest score of performance metrics as accuracy 94.42%, kappa statistics 93.5%, and Avg. TP Rate 94.4%. In addition, we calculated the BOD, COD and DO for one scenario. This study includes prototype hardware details of the proposed IoT system.

# **Table of Contents**

# List of Figures

# List of Tables

# Chapter 1

# Introduction

Natural fisheries limit the number of fish that may be caught and are available in specific months of the year. So, fish farming is very important because of the food supply for humans as well as for fisherman to earn their living. Nowadays, water is being polluted due to various reasons. So it is necessary to monitor the water quality for fish farming. If it is checked by the internet of things (IoT), then it will be helpful for the fish farmers.

## 1.1. Background

Aquaculture means agriculture primarily for the food of aquatic animals or plants or pearls. When it is for only fish farming, then it is called aqua fisheries. There are mainly two types of fish farming. There are many types of aqua fisheries such as extensive fish farming, intensive fish farming, ditch fish farming, and cage system, etc. Extensive fish farming means that it takes place in a pond where intensive fish farming is farming in closed circulation water. In a ditch system, water retaining is the main requirement of this system and it also a crucial point to keep electrolytes for fishes corrected. And cage is a system where it confines the fishes in a mesh enclosure. Overall, aquaculture contains the cultivation, nutrition, and harvesting of fresh and saltwater fish, mollusk, crustaceans, and seedlings. The practice began in China approximately 4,000 years ago and international production remains undermined by China and other Asian states. Aquaculture is used by some of the poor populations around the globe and by major corporations to harvest food. Aquaculture now provides more than half of all human consumption of seafood, a proportion that continues to increase in terms of the global population. Aquaculture produced 3 million tons of food in the seventies, a figure that steadily increased to more than 80 million tons in 2017, as per the Food and Agriculture Organization (FAO) [1]. Fish is among the most useful sustenance on earth. It is stacked with huge enhancements including protein and supplements D. In addition, fish is a rare source of omega-3 healthy fats, which are incredibly huge for the human body and psyche. Fish has loaded down with various enhancements that a large number of individuals are deficient. This consolidates great protein, iodine, and various supplements and minerals. Oily species are often considered the most helpful. That is because of the oily fish. Fat-based improvements are found in greater quantities in salmon, trout, sardines, fish, and mackerel. It fuses supplement D, a fat-dissolvable enhancement that various people are lacking. The oily fishes furthermore gloat unsaturated fats like omega-3 are essential for a healthy body, brain, and unequivocally associated with a diminished threat of various ailments [2]. The two most common causes of sudden death in the world are respiratory disappointments and strokes [3]. A fish is seen as one of the most heart-sound sustenances that we can eat. Clearly, various colossal observational assessments show that people who eat fish regularly get a lesser threat of respiratory disappointments, strokes, and death due to cardiovascular disease [4-7]. The omega-3 fat docosahexaenoic destructive (DHA) is especially huge for the cerebrum and eye progression [8]. It is often recommended that pregnant and breastfeeding women eat enough omega-3 unsaturated fats [9]. Various observational examinations show that people who eat more fish have all the more moderate movements of mental activities [10].

For these significant factors, pisciculture or fish farming is very much important for producing more fish for human consumption. Before cultivating fish, it is mandatory to know which factors affect water quality parameters including ph, turbidity, water depth level, temperature, etc. If the cultivation system depends on the digital system, then it will be more beneficial for the people. The Internet of Things (IoT) system is applied in this system. The IoT characterizes the network of actual items "things" that are inserted with sensors, programming, and selective innovations to interface and supplanting data with remarkable gadgets and frameworks over the Internet[11-12].

The Optimum fish cultivation is subject to the physical, substance, and natural characteristics of water to the vast majority of the degree. Water quality is dictated by factors like temperature, straightforwardness, turbidity, water tone, carbon dioxide, pH, alkalinity, hardness, unionized smelling salts, nitrite, nitrate, essential profitability, microscopic fish populace [13].

The job of different variables like temperature, turbidity, carbon dioxide, pH, alkalinity, hardness, smelling salts, nitrite, nitrate, essential profitability, conductivity, microscopic fish populace, and so on, cannot be disregarded for keeping a sound oceanic climate and for the creation of adequate fish food life forms in lakes for expanding fish creation [14]. Consequently, there is the need to guarantee that, these ecological components are appropriately overseen and managed for good endurance and ideal development of fish. Temperature, pH, turbidity, and conductivity among these factors of water have a much important role in the survival of fish species [15].

Dissolved oxygen (DO) is another important parameter for fish farming. It is an oxygen gas dissolved in water. We have to check the level of it for fish survival. The standard recommended value is 5 mg/l for optimum fish health. The majority of DOs in ponds are generated by aquatic plants and algae during photosynthesis. This is why, DO increases before dawn, decreases at night, and is lowest shortly before dawn. Dissolved oxygen levels below 5 mg/L can be hazardous to fish, whereas surface gulping air can be seen if DO falls below 2 mg/L. An electronic oxygen meter or a chemical test kit may be used to measure dissolved oxygen. When DO falls below 4 mg/L or ambient circumstances promote an oxygen depletion event, emergency aeration should be provided [16]. Another important parameter is biochemical oxygen demand (BOD) for fish farming. Another name is biological oxygen demand (BOD) [17]. It is the quantity of oxygen necessary for the biological breakdown of organic matter in bodies of water. In general, the BOD is a pollution measure used to determine the quality of effluent or wastewater.

The study experiments on the temperature, pH, conductivity, depth, and turbidity parameters of water. In the future, we will work on the rest of the factors.

## 1.2. Internet of Things (IoT)

The Internet of Things is an interrelated method for computers, mechanical and digital machinery, objects, cattle, or persons who are capable of sending data through the network without the need for interaction between human to human or human to computer. IoT tells the physical-object network integrated with sensing, software, and other technologies to connect or interchange data to other devices or systems via the Internet [18]. IoT has a unique identification (UIDs) and can send values over the Internet without the support of human beings. IoT has made progress with the integration of diverse technologies, proper analysis, machine learning, typical science, and systems engineering. In addition, the Internet of Things is enabled in traditional fields such as fixed systems, wireless sensors networks, control systems, automation. IoT is mainly equated with product lines on the customer base that assist one or more common ecosystems and can be managed by ecological system-related devices, such as smartphones including devices and equipment (for example, light fittings, regulators, house safekeeping systems and cameras, and other domestic devices). IoT can also be used in the healthcare sector, agriculture sector, smart cities, etc [19-20]. Figure 1.1 displays the application of IoT in various sectors.

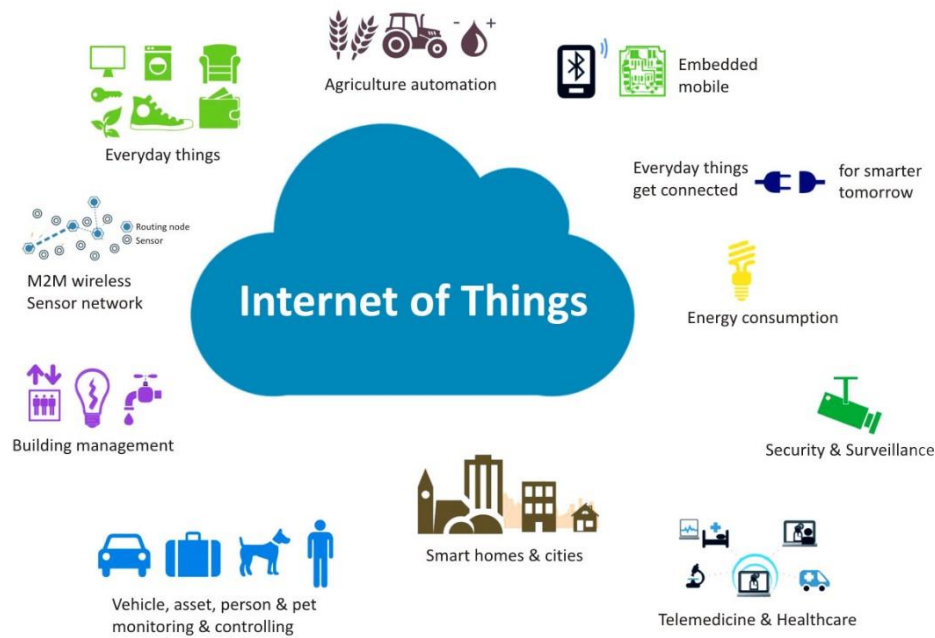

Figure 1.1: IoT applications [21]

## 1.3. Objectives of the Thesis

The objective of the thesis is mentioned below:
- To create an IoT framework for real-time data visualization
- To check the suitability of a pond for fish farming according to the standard reference values of various aquatic parameters
- To analysis the water quality parameters
- To generate real-time values and apply machine learning models for fish prediction

## 1.4. Contribution of the Thesis

The following is the thesis' main research contribution:

- A real-time value dataset
- A prediction model for fish classification using a machine learning model
- Fish farmer can understand about the ponds; what ponds are suitable for fish farming and what are not
- Used machine learning model for data analysis

## 1.5. Overview of the Thesis

Water is very important on the earth. It is essential for the survival of all lives. In the thesis, we experimented with water of ponds for fish farming. There are many elements or parameters in water including temperature, pH, turbidity, conductivity, biochemical oxygen demand (BOD), dissolved oxygen (DO), etc. These parameters are important for fish survival. In the thesis, we considered 5 ponds for cultivating fish. First of all, we set up a system using various IoT devices and Arduino and ethernet shield. Using this system, we collected the real-time values of some parameters including pH, temperature, turbidity. We also collected conductivity, depth values where we got conductivity by using an established method and we got the value of depth by using sticks. After that, we compared these real values with the standard reference values which we got from the various research paper. Finally, after comparison, we can decide that what ponds are satisfied and what are not.

In addition, after labeling fish categories with real values, we did analysis these values using various machine learning models including Random Forest, K-NN, J48, and so on. Among them, Random Forest provides the best performance metrics.

## 1.6. Outline of the Dissertation

Chapter 2 describes the real-time monitoring of water quality parameters and IoT system literature evaluation and different types of sensors. Chapter 3 represents the proposed methodology, physical system implementation, and the cloud server of the thesis. In this chapter, the experimental setup is demonstrated also. Chapter 4 gives a detailed description of data analysis using the machine learning models. Chapter 5 describes the discussion and result of the analysis. Chapter 6 tries to give a proper finishing with a conclusion of the whole thesis and future work of the thesis towards further research.

# Chapter 2

# Literature Review

According to our thesis title and after understanding the objective and outcome of the thesis, some keywords have been fixed up. By using these keywords, several research papers have been searched through different search engines. Among all the research papers, some of them are related to our thesis work. This chapter also describes existing key methodologies, related tasks, and hardware devices briefly.

## 2.1. Related Works

This section discusses a segment of actions linked to the embedded system using IoT for pisciculture. Many IoT-based water inspection frameworks have been proposed as of late for various territories of utilizations for river and seawater checking [22-25]. Some of them are effectively utilized for fish homesteads and hydroponics focuses [26]. An IoT system is proposed for selecting the fish species using various sensor values including mq7, pH, turbidity, ultrasonic, and temperature sensors [27]. The main limitation of this paper is not describing the real-time scenarios of the pond's environments. In [28], the authors proposed an IoT system for fish farming thoroughly. But, no real-time values and scenarios are generated for fish farmers. A water quality monitoring system based on the IoT is proposed in the aquaculture in Mekong Delta [29]. The narrowness of this system is to cultivate one fish type called pangasius at five farms. WSN (Wireless Sensor Network) based monitoring system has been proposed in paper [30] for water quality and fish behavior during the feeding process in aquaculture. It is designed for tank water only. In [31], an IoT system is mentioned for fish farming. The restriction of this system that it is designed for only fish named guppy fish and only two sensors are used including pH and salinity sensor. In [32], the authors proposed a data-driven approach for monitoring the aquatic environment for the ocean. The limitation of the model, it creates a stochastic model for predicting the actual ocean data without using IoT devices. In [33], the authors discussed the parameters only of aquatic environments without any IoT devices. A secure fish farm system is raised in tank water using blockchain technology without IoT concept [34]. In [35], a remote monitoring system is proposed using the IoT. An IoT system has been proposed using only two sensors for fish farming in Bangladesh [36]. It is not appropriate for detecting all quality factors of the water. In [37], temperature, pH, DO levels, ultrasonic sensors are used for monitoring the tank water. The narrowness of this system is that it is conducted without any real-time scenarios. A mobile app for monitoring water quality is proposed in [38], where two sensors including pH and temperature are used. The limitation of the model is that it is not suitable for measuring all factors of water. Although the state-of-art models show models for aquatic environment monitoring systems, very few papers deal with real-time environment monitoring for cultivating fish species in a pond. In [39], an IoT-based system for fish feeding is designed using a pH sensor and ultrasonic sensor. A system is developed for observing aquaculture using dissolved oxygen and temperature sensor based on a wireless sensor network [40]. A mobile application is designed for fish feeding using the internet of technology [41]. An IoT system is built for water pumps in fish farming using only one sensor named water level [42].

In [43], a structure is built which is showing real-time indicators for aquatic ecosystems, and transactional data, and assessing the impact on fish production species survival and biofuels, along with production failures. In [44], the authors showed the prediction using one feature of water called dissolved oxygen (DO) for an aquatic creature. A device is made for monitoring water quality factors including pH, temperature, and dissolved oxygen [45]. An IoT device is proposed for detecting and controlling water factors including pH, temperature. But they did not analyze the data [46]. A regression model is utilized for predicting water quality for cultivating fish [47]. But they did not show any accuracy. In [48], the authors proposed an automated strategy for fish identification primarily based on the support vector machine and k-means clustering algorithms. Another computerized Nile Tilapia fish classification approach was proposed in [49]. The authors used scale-invariant characteristics to seriously change and study elements for characteristic extraction. Then, these points are used to feed the support vector machine. The authors mentioned the management of aquaculture production through psychical, biochemical process rules and calculation [50]. A scientific model is designed to assess ecological influence [51]. In [52], the authors developed ML algorithms for predicting the spread of algae along the watercourse. The ideal range of utilized parameters is shown in Table 2.1.

TABLE 2.1 Ideal range of utilized parameter of water

| Parameter | Ideal range | Findings |
|---|---|---|
| pH value | 6.5-8.5 | [53] |
| Temperature Value | 16-24$^O$C | [54] |
| Turbidity value | Below 10 ntu | [55] |
| Conductivity | 970-1825 μS/cm | [56] |
| Depth | The pond should not be less than 1 meter or more than 5 meters, the best depth is 2 meters. | [57] |

To overcome the limitations of the state-of-art models, in this thesis, we propose an IoT framework for real-time aquatic environment monitoring using Arduino and sensors and a dynamic website for end-users. To evaluate the proposed architecture, we have utilized five ponds of various sizes and specifications.

## 2.2. Overview of Arduino

Arduino is an open-source prototype framework with simple hardware and software. It contains a programable circuit board and ready-to-use software known as the Arduino IDE which can be used to create and send programming code to the Arduino platform.

The distinguishing characteristics are as follows:
- Arduino boards can read analog or digital input signals from a variety of sensing devices and convert them to a result
- Arduino IDE can be used to monitor the functions of the board by transferring a series of directions to the board's microcontroller

- Additionally, the Arduino IDE makes programming simpler by using a simplified version of C++
- At last, Arduino offers a regular user interface that condenses the microcontroller's capabilities into a more manageable kit.

### 2.2.1. Arduino Uno

In contrast to the ATmega328P, the Arduino Uno refers to a microcontroller board that has 14 modernized data/yield pins, a sixteen mega Hartz quartz crystal, a universal serial board link, a jack power, in-circuit serial programming, and a reset pin. It comes with all where the microcontroller needs to start-up by connecting to a PC through USB or an AC-to-DC adapter.

### 2.2.2. Technical Specification of An Arduino Uno

There are some specification parameters of the Arduino controller. They are- operating voltage for 5V, input voltage (recommend) for 7-12V, input voltage (limit) for 6-20, digital total pin for 14, analog input pin for 6, DC per I/O for 20mA, DC for 3.3V pin for 50mA, flash memory for 32KB, static RAM for 2KB, electrically erasable PROM for 1KB, clock speed for 16 MHz, length for 68.6 mm, width for 53.4 mm, etc.

### 2.2.3. Arduino Power

The Arduino Uno can be powered by a USB link or an external power supply. The control source is mechanically picked. The external power can be provided through an AC-to-DC adaptor or a battery. The power pins are as follows:

- VIN: When using an external power source, this is the information voltage to the Arduino board. Voltage through this pin can be supplied

- 5V: The microcontroller and various segments on the board are operated by a regulated control supply. This can be provided by a USB link or regulated 5V power supply.
- GND: Ground pin

### 2.2.4. Arduino Memory

The memory management of ATmega 328 is 32 KB of flash memory, 2 KB of static RAM, and 1 KB of an electrically erasable PROM for storing code.

Arduino input and output:

The Arduino I/O pins are shown graphically in Figure 2.1.

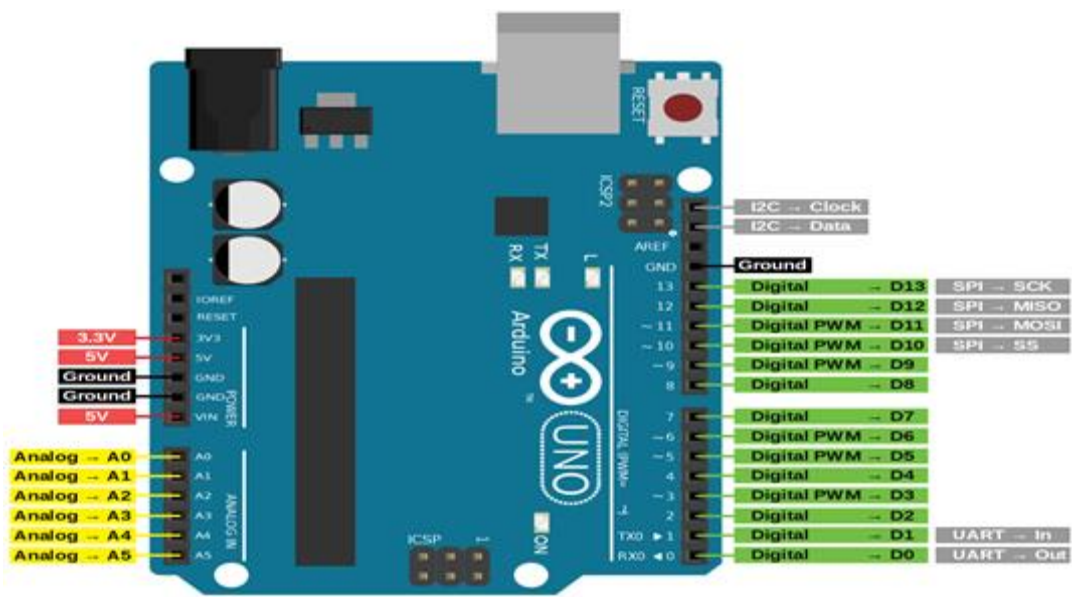

Figure 2.1: Arduino I/O pins [58]

Description:
- The range of analog pin is A0-A5
- The upper in of 13 is digital ground
- Digital pin range is 2-13
- Pins in digital are 0-1/Serial In/Out - TX/RX (dark green)
- S1 is reset button

- Total three ground pins
- USB port is used for connecting to the computer

Every pin name is written on the board.

### 2.2.5. Arduino LED

The Arduino has four LEDs: L, RX, TX, and ON. On the board, the ON button is on the bottom right, and the L, TX, and RX buttons are on the upper right corner of the Arduino Uno.

- ON Lead: When the Arduino is powered up, this LED will flash orange. If our Arduino isn't acting properly, keep an eye on this LED; if it's gleaming or off, it's time to check our power supply
- RX and TX LED: These are identical to the send and receive LEDs on the connection modem. They flash when data is sent from or not sent to the Arduino through the USB connection. Whenever values can be transferred after Arduino to PC USB port, the TX LED illuminates yellow and RX LED illuminates will be yellow vise versa
- LED: This is the only LED to have power over. The ON, RX, and TX LEDs all light up naturally no matter what; however, the L Drove is connected to the Arduino principle chip, and we can turn it on or off when we start writing code and transferring on it

## 2.3. Ethernet Shield

Ethernet shield allows easily connecting Arduino Uno to the Internet. This shield enables the Arduino to send data to a cloud server and to receive data from sensors with an Internet connection. The Wiznet W5100 ethernet chip is used in this shield. This chip contains the PHY, MAC, IP, and TCP layers. Hardware implementation on the chip is the merits of using the shield through the ENC28J60. Transfer controlled protocol or internet protocol must be executed on the microcontroller where it is connected to the ENC28J60 chip.

The shield also includes a microSD card socket for connecting to a microSD card. The SPI bus is used by Arduino to communicate with both the W5100 and the microSD card. This is connected to 13-10 pins, and pin 4. It is easy to program with the shield that Arduino library is available for both Ethernet and SD card.

Ethernet Shield Features:

- Operating voltage is 5 volts
- Input voltage is between 7 to 12 volts
- ATMega328.
- Digital I/O pins are 14
- Analog Input pins are 6
- Available micro SD card slot

Ethernet Shield LEDs:

- PWR: Demonstrates that the board and shield can be controlled
- LINK: Shows the nearness of a system connection and flashes when the shield transmits gets information.
- FULL D: Demonstrates that the system association is double-faced
- 100M: Shows the nearness of a 100Mb/s arrange association (instead of 10 Mb/s).
- RX: When the shield receives information, it flashes.
- TX: When the shield sends information, it flashes.
- COLL: When an organized impact is detected, it flashes.

## 2.4. Survey of Sensors

The sensors used in the implementation are discussed in the following sections. They are- pH sensor, temperature sensor, turbidity sensor, and ultrasonic sensor. We used this in the thesis for measuring the real-time values from the various 5 ponds.

## 2.4.1. pH Sensor

The potential of hydrogen (pH) meter is a systematic tool that deals with the hydrogen-ion movement in aquatic-based solutions to determine their tartness or alkalinity, which is said as pH. A pH meter is made up of two basic components: a pointer that moves against a scale and a digital meter that takes value from the resources and displays it numerically via a circuit board. In our thesis, we have a pH sensor which is a digital meter that we use to test the acidity of the water. We build a circuit board and connect it to an Arduino. There is some code in Arduino that works with a pH meter. The meter has a 14 scale from 0 to 14, and it calculates the acidic or alkalinity of a solution. It is deemed neutral when its value is 7. When above 7, the solution is alkaline, and below 7, then acidic water is considered. The ideal range of pH for fish farming in a pond is between 6.5 and 8.5. If the pH value is below 4 and above 11, then it is the death point for fish due to the acidity and alkalinity. When the range of pH is 4-5, no reproduction of fish occurs, and when it is in the range of 4-6.5, and 8.5-10, slow growth will be for fish. So it is the major survival element of water.

It may be properly quantified using the sensor which can calculate the potential variance between two electrodes: a reference electrode and a glass electrode sensitive to hydrogen ion. Overall, the pH sensor can be used easily and provides more accurate calculations. It is also reusable. The probe's procedure is as follows.

**Principle of pH Meter or pH Sensor:**

The pH meter is based on the fact that the interface of two liquids generates an electric potential that can be measured. In other words, when a liquid inside a glass enclosure is placed inside a solution other than that liquid, an electrochemical potential exists between the two liquids.

**pH Sensor Components:**

The electrode of pH sensor has four parts. They are-
- Glass tube: A glass tube called measuring electrode with a thin glass bulb soldered to it which is covered with a potassium chloride solution. This also has a silver chloride block where it produces the voltage required to determine the pH of an unfamiliar solution.
- A Reference electrode: This is also a glass tube containing a potassium chloride solution which is used to complete the circuit by providing a stable zero-voltage connection.
- Preamplifier: A signal training device that transfers the high resistance pH signal to a low resistance signal. It is also used for the signal's strength and stability.
- Transmitter: This part displays the electrical signal of sensors and includes a temperature sensor to recoup for values changes.

The pH value of daily used substances including lemon juice as 2.40-2.6, cola drink as 2.5, vinegar as 2.5-2.9, orange juice as 3.5, beer as 4.5, coffee as 5, tee as 5.5, milk as 6.5, water as 7, saliva as 6.5-7.42, blood as 7.38-7.42, seawater as 8, soap as 9-10, and bleach as 13.

### 2.4.2. Temperature Sensor

The single wire protocol digital temperature sensor, such as the DS18B20 model can calculate the temperature on the scale of -55ºC to +125ºC with ±5 percent precision. The data established from the 1-wire is in the 9-bit to 12-bit range. This sensor may be controlled by a single pin on a microcontroller since it follows the single wire protocol. The sensor can be programmed with a 64-bit serial code for the progressive level protocol, allowing multiple sensors to be controlled from a single microcontroller pin.

There is 3 colored pin named black color, red color, and yellow color. A black pin is used to connect to the GND. The red pin is known as vcc which varies from 3.3V or 5V. Yellow pin supplies the output.

### 2.4.3. Ultrasonic Sensor

Ultrasonic sensors are excellent instruments for measuring distance without making physical contact, and they are used in a variety of applications such as water level measurement, distance measurement, and so on. This is a quick and accurate way to calculate small distances. We used this sensor to assess the distance of a hindrance from the sensor in the thesis. The ECHO principle is the foundation of ultrasonic distance measurement. When sound waves are transmitted in the area, they return to their source as ECHO after colliding with an obstacle. So, after striking the barrier, we just need to measure the travel time of both sounds, i.e., the outgoing time and the return time to origin. Since we know the speed of sound, we can measure the distance with a little math.

This type of HC-SR04 is a sensor that is primarily handled to assess the distance between the goal object and the sensor. It uses non-contact technology, which means there is no direct contact between the sensor and the object being measured.

Transmitter and receiver are the two main components with the previous converting electrical signals to ultrasonic waves and the latter converting ultrasonic waves back to electrical signals.

It provides detailed dimension information and has a resolution of about 3mm, implying that the actual distance between the item and the measured distance can vary slightly. There are 4 pins of the sensor including VCC, trig, echo, and ground pin where VCC is used for power supply at 5 voltage, the trig pin is used to initialize calculation for transmitting ultrasonic waves, echo is used for the bounce-back of ultrasonic waves, and ground pin is connected for ground.

### 2.4.4. Turbidity Sensor

Turbidity is a measurement of how many suspended particles are present in a stream. It may be soil in our water or chocolate chips in our milkshake. Although we all want chocolate in our drinks, soil particles are completely unwelcome. Aside from potable uses, Water is employed in a wide range of industrial and domestic settings, for example, Water is used to clean the windshield of a vehicle, it is used to cool the reactors of a power plant, and washing machines and dishwashers rely on it like fish.

There are some parameters of turbidity sensor including working voltage for DC 5volt, operating current for 30mA maximum, response time for less than 500ms, resistance for 100M minimum, analog output for 0-4.5, the weight for 55g, etc.

## 2.5. Connection with Arduino

In this section, how to connect all sensors to Arduino and Ethernet shield is displayed here to measure the real-time values of the aquatic parameters in the ponds.

### 2.5.1. DS18B20 Temperature sensor connection with Arduino

The DS18B20 Digital Thermometer offers temperature readings that are 9 to 12-bit and shows the device's temperature. The DS18B20 communicates with a central microprocessor through a 1-Wire bus that needs only 1 data line by definition. Furthermore, the DS18B20 may draw power directly from the data line, obviating the use of an additional power supply depicted in Figure 2.2.

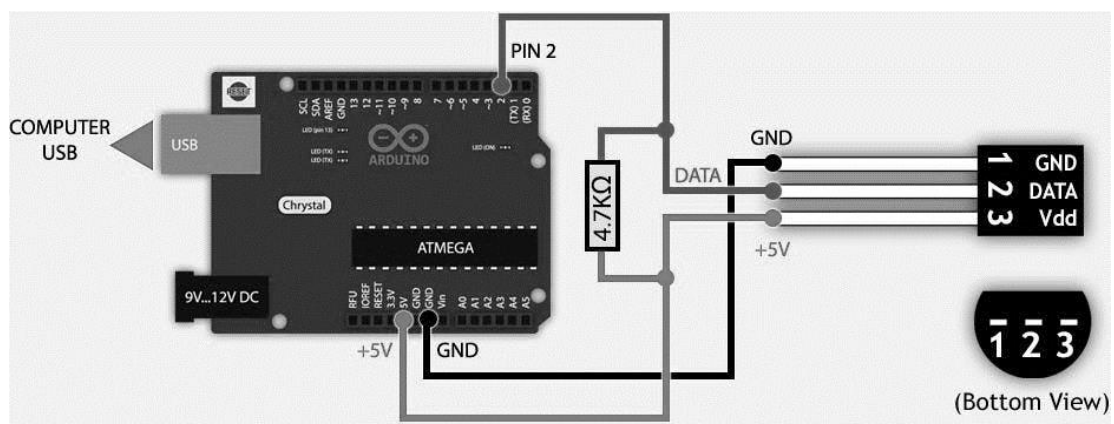

Figure 2.2: Temperature sensor connection with Arduino [59]

Another feature of the DS18B20 is its ability to function without an external power source. When the bus is overloaded, power is therefore delivered via the DQ pin through the one wire pull-up resistor. When the bus is higher, an internal condenser is stored and feeds the system with power when the bus is low. The phrase "parasitic control" applies to the single wire bus system. The DS18B20 can also be operated from a VDD-connected external supply.

### 2.5.2. PH sensor connection with Arduino

We follow the following steps to interface the pH kit with Arduino shown in Figure 2.3. Three wires are available in the sensor. The red wire is connected with the Arduino's VCC and the black wire is connected with the ground, finally, the blue wire is connected to analog pin A0.

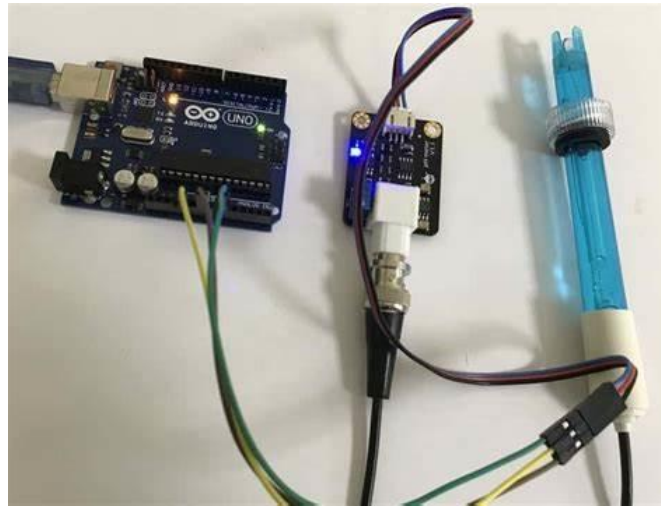

Figure 2.3: Ph sensor connection with Arduino [60]

### 2.5.3. Ultrasonic sensor connection with Arduino

It has some metrics for technical specifications. They are- power supply, current, resolution, measuring angle, distance scale, etc. 5 volts are taken for power supply, less than 2mA for quiescent current, and 15mA is supplied for current. 2cm -400cm is the varying distance and 30 degrees is the measuring angle. Figure 2.4 shows the Arduino and ultrasonic sensor circuit diagram for measuring distance. The "trig" and "echo" pins on the ultrasonic sensor module are directly linked to Arduino pins A4 and A5 in the circuit. In 4-bit mode, an Arduino is connected to a 16x2 LCD.

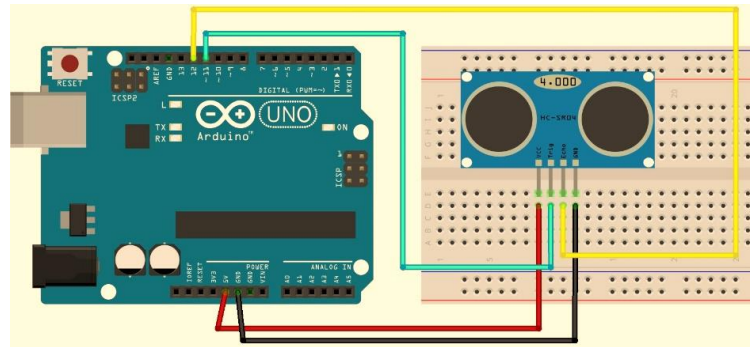

Figure 2.4: Ultrasonic sensor connection with Arduino [61]

The Arduino pins 2, GND, and 3 are directly attached to the control pins RS, RW, and En. Data pins D4-D7 are wired to Arduino pins 4, 5, 6, and 7. First, we must use Arduino to activate the ultrasonic sensor module to send a signal, and then we must wait for the ECHO to arrive. The time between triggering and Received ECHO is measured by Arduino. The formula for the speed of sound is about 340 meters per second. As a result, we can measure distance by using equation 2.1.

$$Distance = \left(\frac{travel\ time}{2}\right) \times speed\ of\ sound \qquad (2.1)$$

Where the speed of sound around 340m per second.

### 2.5.3. Turbidity sensor connection with Arduino

The sensor is to be connected to Arduino using three pins only named- VCC, GND, and signal. Figure 2.4 shows the turbidity sensor is connected to an Arduino Uno.

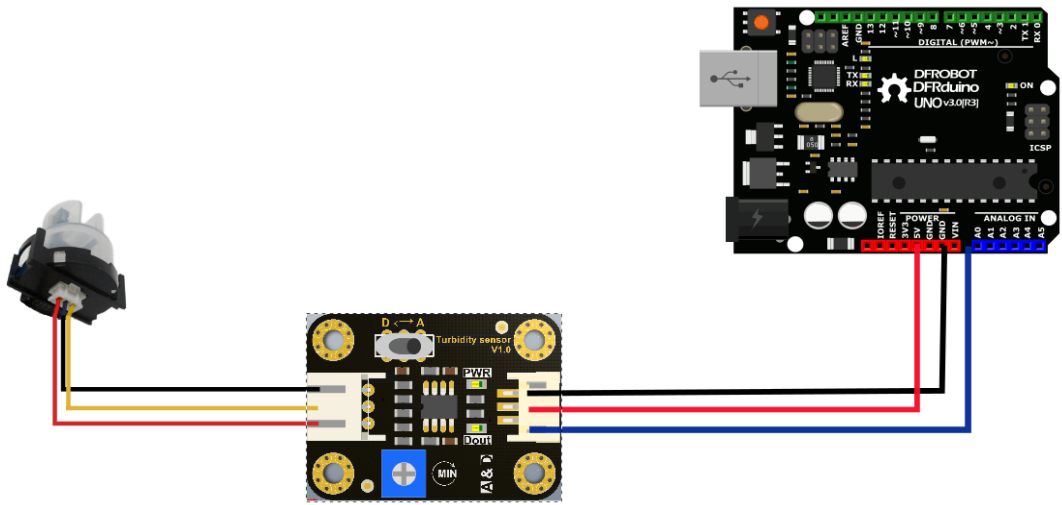

Figure 2.4: Turbidity sensor connection with Arduino [62]

After a rigorous review of the above-related research materials, we understand the different methodology, different tasks, and explanations of their respective hardware devices. We also got a different limitation of existing work. Moreover, we find out review materials about the different parameters of the aquatic environment which helps us to determine the range of different parameter values for further taking decision processes. The following chapter will demonstrate the proposed methodology.

# Chapter 3

# Proposed Methodology

In this chapter, the proposed methodology is discussed. The physical implementation of the system, the experimental setup for getting the real-time values of the water quality parameters, IoT cloud server are demonstrated.

## 3.1. Overview of the Methodology

Figure 3.1 presents the proposed methodology. First of all, we introduce all the aquatic sensors such as pH sensor, temperature sensor, turbidity sensor, etc. Every sensor is connected with an Arduino Uno through an ethernet shield by a various jumper on the breadboard. Using the Rest-API, a cloud server is connected with this system.

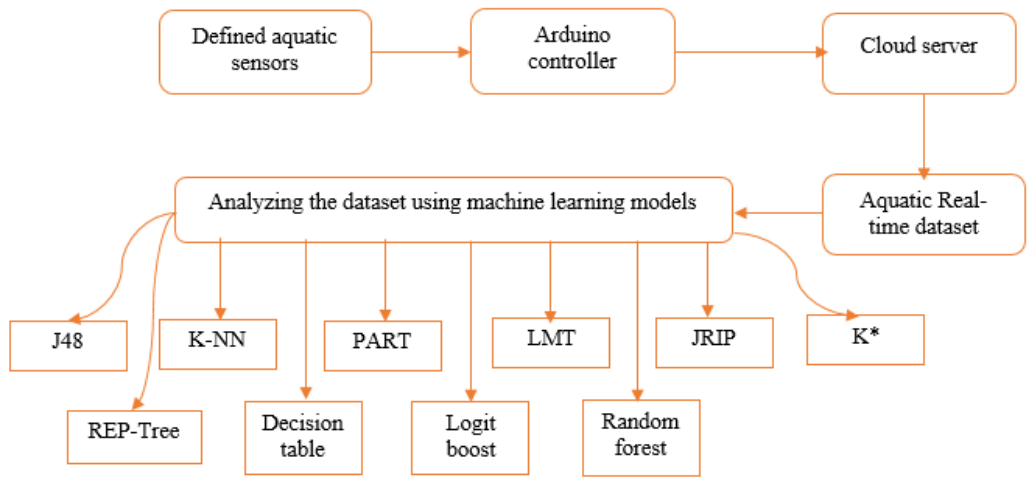

Figure 3.1: Illustration of the Suggested methodology

Then, we use 10 machine learning models including J48, K-NN, random forest, K*, LMT, PART, JRIP, decision table, logitboost, and REP-Tree for the fish classification.

## 3.2. IoT Framework Implementation

In the IoT system, We used 4 sensors for measuring real-time data of each pond water. They are as follows: a ph sensor, a temperature sensor, a turbidity sensor, and an ultrasonic sensor.

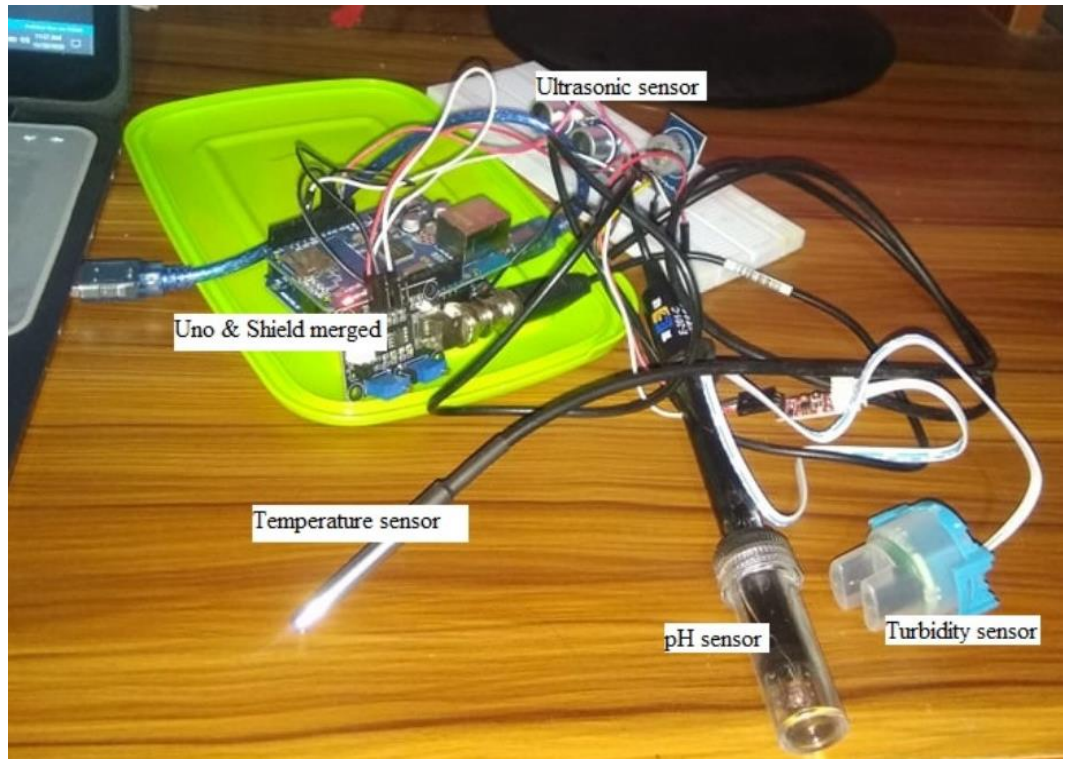

Figure 3.2: Physical implementation of the proposed IoT system.

Figure 3.2 shows a depiction of all hardware utilized in the proposed IoT framework. We used a temperature sensor for controlling temperature. It is significant for fish cultivation as it can influence their conduct, taking care of, development, and propagation. The pH is an assessment that contacts the acidity or alkalinity of the solution. This is also named the hydrogen ion concentration guide. It is a scale of hydrogen ion activity in solution. It is numbered between 0 to 14. When its value is 7, then it is called neutral. If it is above 7, then the solution is alkaline and below 7, then it is said acidic water. Generally, the range of pH is 6.5-8.5 for fish survival in a pond. It is the main factor for fish survival. That is why we use this sensor in our system. The ultrasonic sensor can detect the depth level of water. Every fish species cannot live on one level. Fishes live in upper level or middle level or lower level. For measuring this point, we used this sensor in our system. Turbidity is the unwanted something in the water that hinders the production of fish species. For measuring this issue, we used a turbidity sensor. Turbidity brought about by dirt or soil particles can confine light entrance and cut off photosynthesis. Sedimentation of soil particles may likewise cover fish eggs and annihilate advantageous networks of base living beings, for example, microorganisms.

We have utilized an Ethernet shield and Arduino Uno in the proposed architecture. For linking Arduino Uno to the Internet, an Ethernet shield is needed. This shield enables the Arduino to send data to a cloud server and receive data from sensors with an internet connection. The used microcontroller board in the perception of the ATmega328P can identify the climate by getting involvement from a collection of sensors and managing lights, engines and various actuators can alter their environmental impacts. The microcontroller is customized utilizing the Arduino programming language and the Arduino advancement climate.

### 3.2.1. Hardware Devices

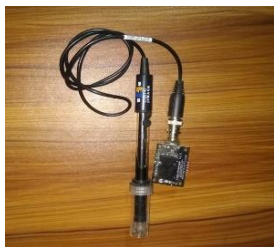 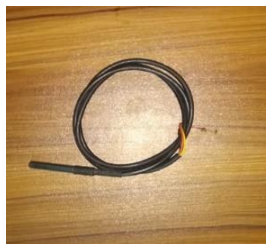 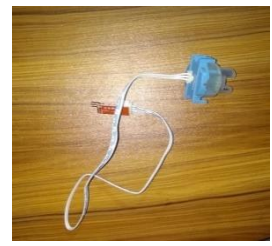

(a) pH sensor      (b) Temperature sensor      (c) Turbidity sensor

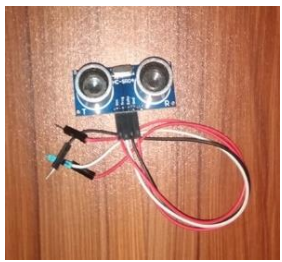 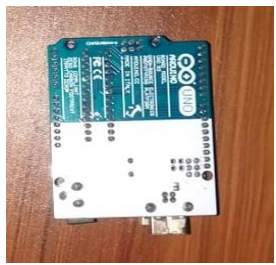 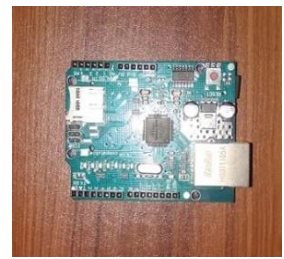

(d) Ultrasonic sensor      (e) Arduino Uno      (f) Ethernet shield

Figure 3.3: Hardware devices

Figure 3.3 shows the hardware devices which is used for building the IoT system and for measuring the real-time values of the aquatic environmental parameters. We see (a) pH sensor, (b) temperature sensor, (c) turbidity sensor, (d) ultrasonic sensor, (e) Arduino Uno, and (f) Ethernet shield from figure 3.3.

## 3.3. Experimental Setup

We have considered 5 ponds for monitoring the quality of water using the proposed IoT system.

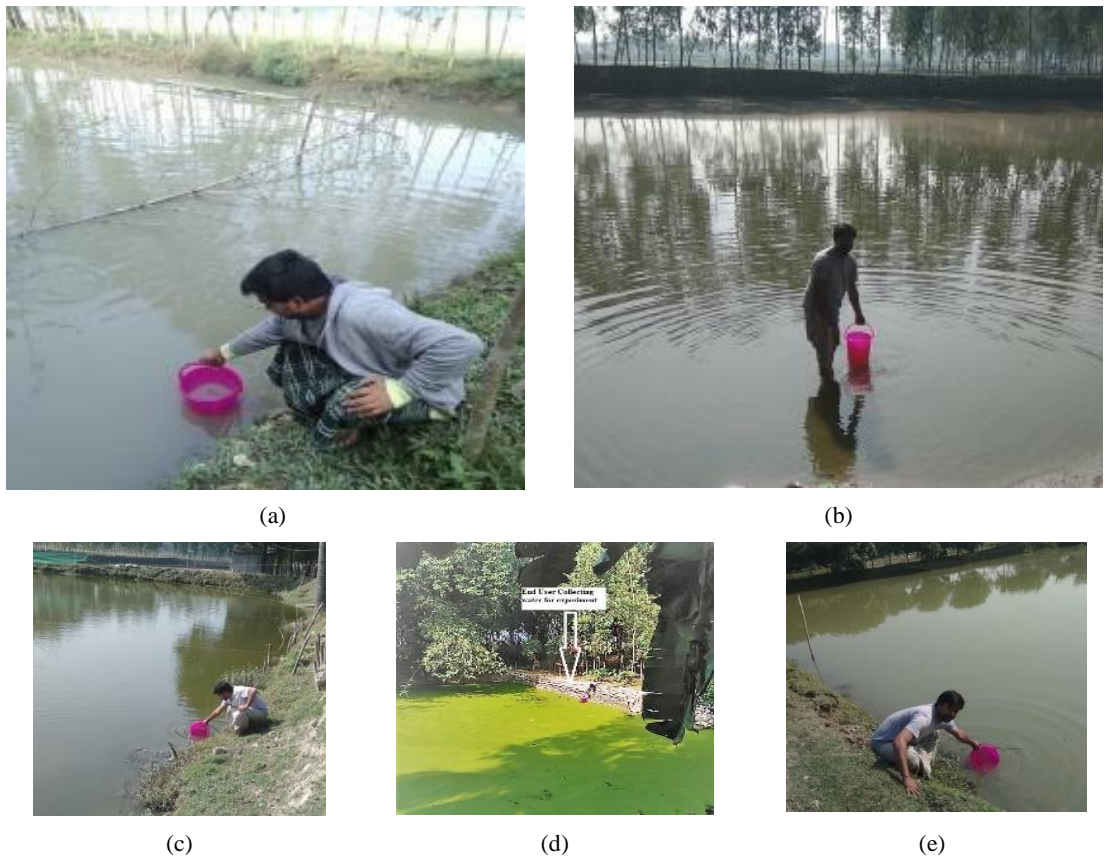

Figure 3.4: Water collection from all ponds (a) Pond 1, (b) Pond 2, (c) Pond 3, (d) Pond 4, (e) Pond 5

In the experimental evaluation, we do programming in Arduino suite software. Due to the lack of portability of the hardware, we have collected water using a bucket from all the ponds, which is shown in Figure 3.4.

Table 3.1 illustrates the specifications of all ponds which is the smallest pond among the 5 ponds.

TABLE 3.1  Specification of all ponds

| Term | Pond 1 | Pond 2 | Pond 3 | Pond 4 | Pond 5 |
|---|---|---|---|---|---|
| Length(m) | 26 | 52 | 105 | 156 | 40 |
| Width(m) | 17 | 30 | 35 | 80 | 20 |
| Depth (m) | 1-2 | 1-2 | 1-2 | 2-4 | 1-3 |

After collecting, we dipped our all sensors into the water for getting real-time data. Figures 3.5, 3.6, and 3.7 show the experiment of pH values, turbidity values, and temperature values, respectively in water for analysis of water quality.

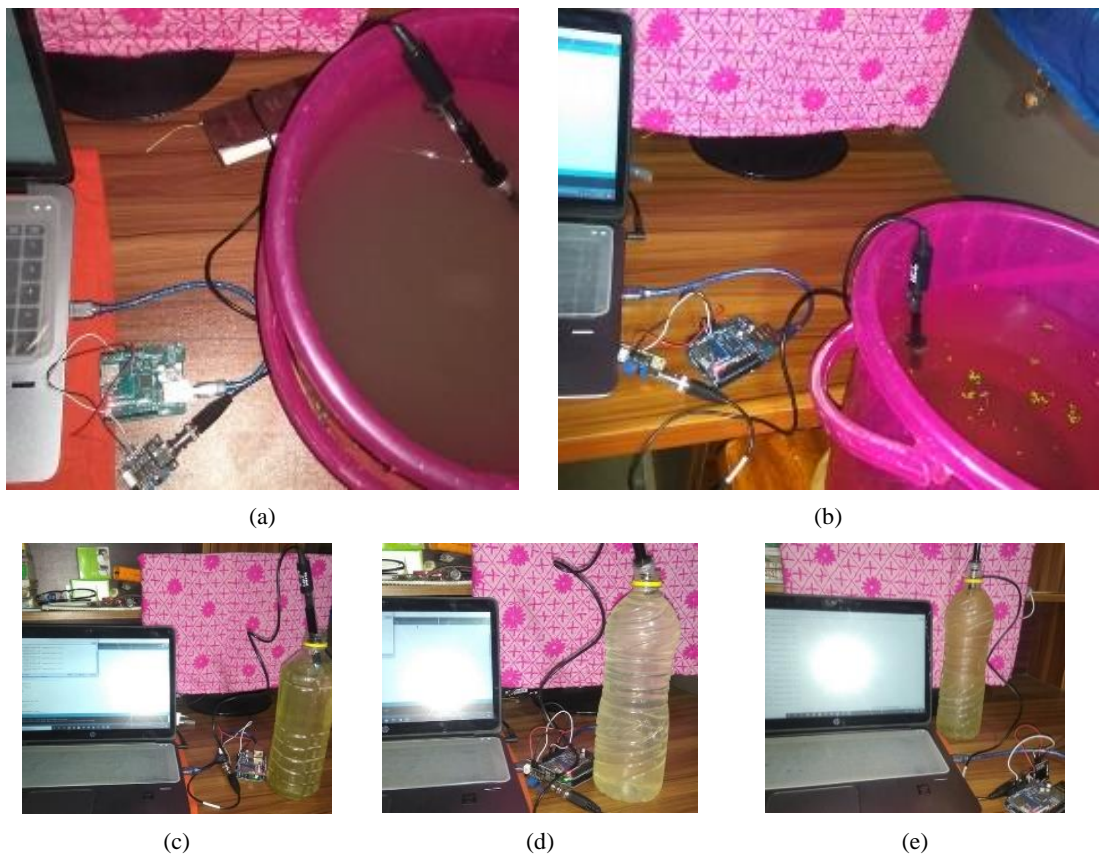

(a)　　　　　　　　　　　　　　(b)

(c)　　　　　　(d)　　　　　　(e)

Figure 3.5: Experimenting water of pH value in all ponds (a) Pond 1, (b) Pond 2, (c) Pond 3, (d) Pond 4, (e) Pond 5.

The experiment of pH sensor of all experiments is between 8:50-9:30 pm at 20 December 2020, on between 11:18-11:50 am at 21 December 2020, on between

2:00-2:50 pm at 20 December 2020, on between 6:00-6:50 pm at 20 December 2020, and on between 2:30-3:20 pm at 22 December 2020 for pond-1, pond-2, pond-3, pond-4, and pond-5, respectively. We take 4 hours 10 minutes from 3 days for pH experiment in all experiments. In the Arduino suite, we keep the baud rate 9600 for serial begin function for all experiments.

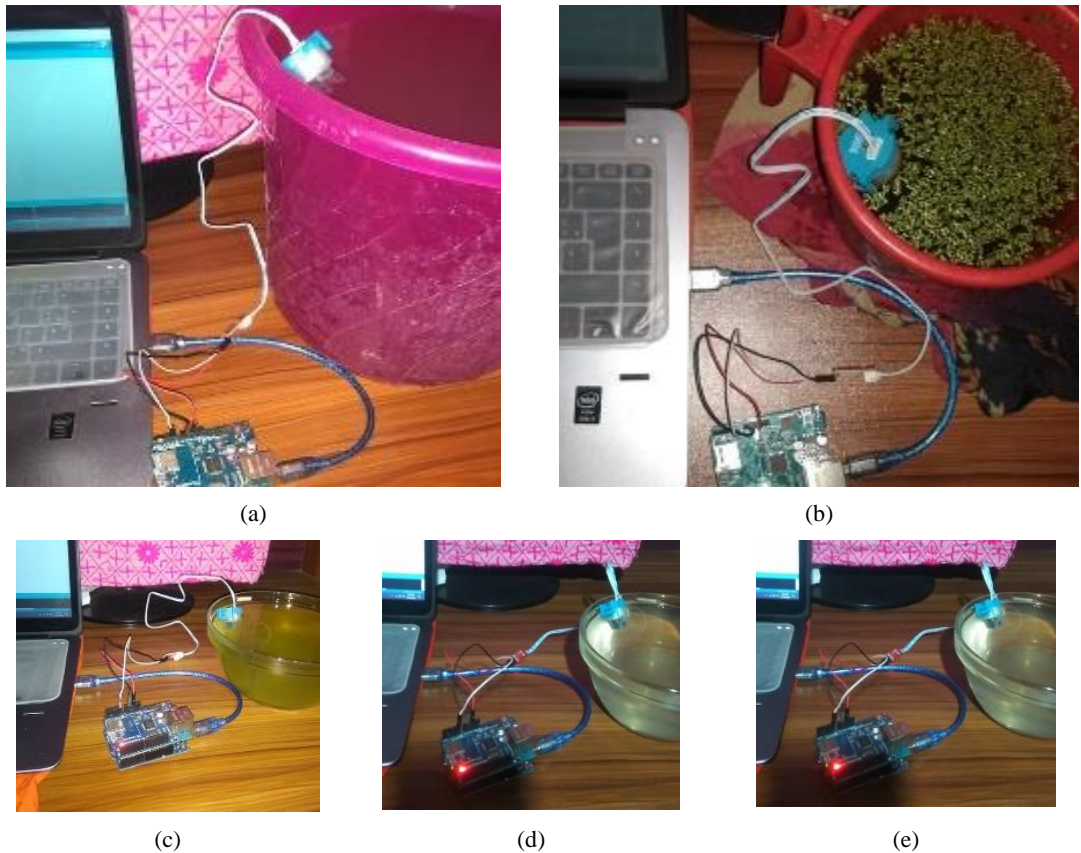

Figure 3.6: Experiment water of turbidity value in all ponds (a) Pond 1, (b) Pond 2, (c) Pond 3, (d) Pond 4, (e) Pond 5.

The experiment of the turbidity sensor of all experiments is on between 10:00 -10:50 am at 23 December 2020, between 12:00-12:50 pm at 23 December 2020, between 2:10-3:00 pm at 23 December 2020, between 4:00-4:50 pm at 23 December 2020, and between 5:20-6:10 pm at 23 December 2020 for pond-1, pond-2, pond-3, pond-4, and pond-5, respectively. We take 4 hours 10 minutes from 1 day for turbidity experiments in all experiments.

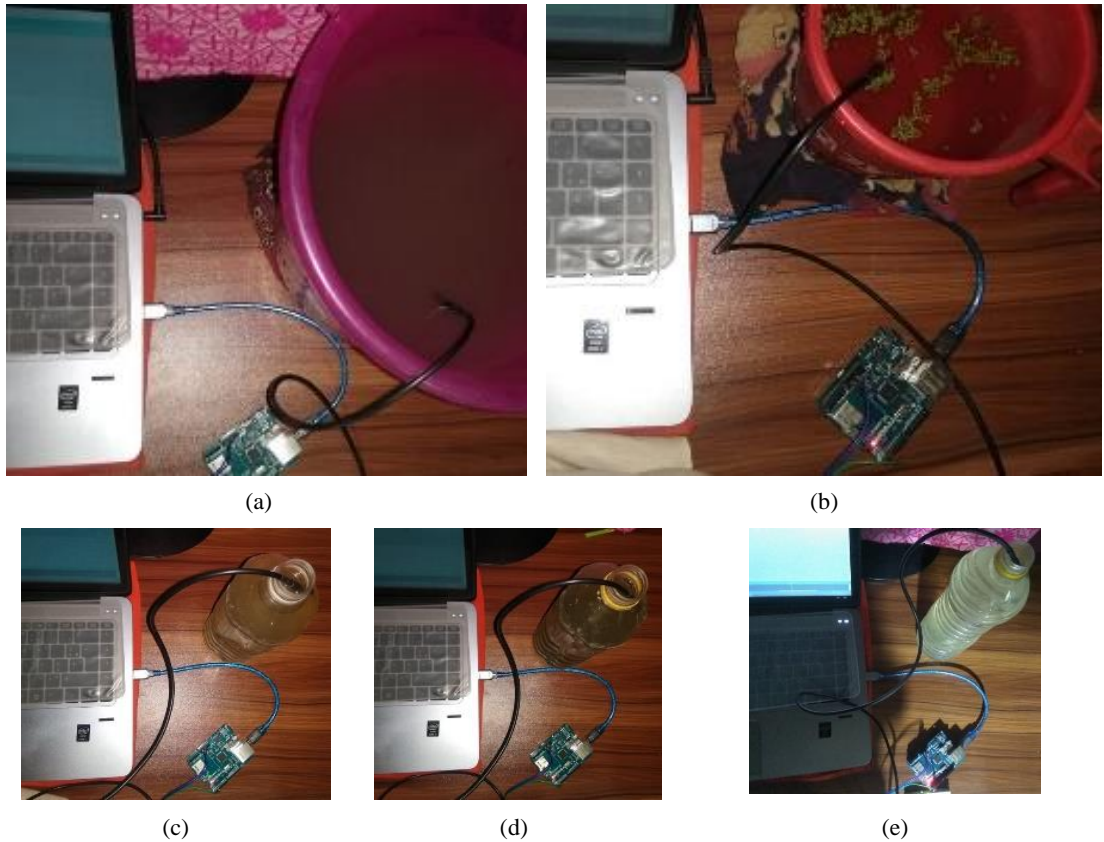

Figure 3.7: Experiment water of temperature value in all ponds (a) Pond 1, (b) Pond 2, (c) Pond 3, (d) Pond 4, (e) Pond 5.

The experiment of the temperature sensor of all experiments is on between 12:00-12:50 pm on 25 December 2020, on between 01:10-2:00 pm on 25 December 2020, on between 02:10-03:00 pm on 25 December 2020, on between 11:09-11:59 am at 26 December 2020, and on between 12:20-1:10 pm at 26 December 2020 for pond-1, pond-2, pond-3, pond-4, and pond-5, respectively. We take 4 hours 10 minutes from 1 day for temperature experiments in all experiments. In the Arduino suite, we keep the baud rate 9600 for serial begin function for all experiments.

After that, the real-time data is stored in a cloud server named ThingSpeak by using REST-API.

## 3.4. Cloud Server

The real-time data which we got from the experimental setup is stored in a cloud server named thingspeak IoT server [63] using rest-API. ThingSpeak is an (IoT) stage that allows us to dissect and imagine the information in MATLAB except purchasing a permit from Mathworks. It permits us to gather and store sensor information in the cloud and create IoT applications. It communicates sensor data to ThingSpeak via Arduino, ESP8266 Wifi Module, Particle Photon and Electron, BeagleBone Black, mobile and web applications, Raspberry Pi, Twilio, Twitter, and Matlab. The ThingSpeak is generally centered around sensor logging, area following, triggers and cautions, and examination. We create a channel named Fish Farm Monitoring System shown in Figure 3.8.

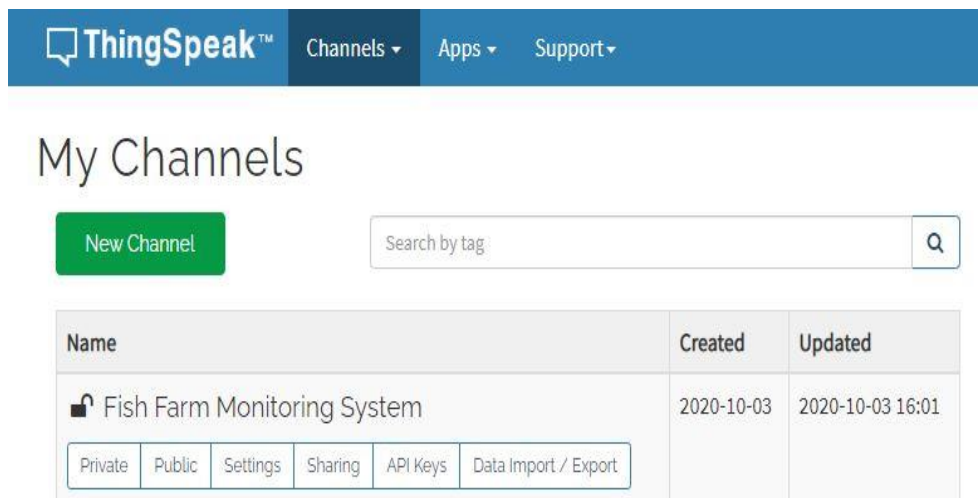

Figure 3.8: Channel window

### 3.4.1. Real-Time Data

The real-time data is stored in the thingSpeak channel. It is stored in CSV, XML, and JSON format, as depicted in Figure 3.9.

![Export recent data table showing JSON XML CSV export options for Fish Farm Monitoring System Channel Feed, Field 1 Data: Turbidity, Field 2 Data: Temparature, Field 3 Data: PH, Field 4 Data: Depth]

Figure 3.9: Real-time data stored in the cloud.

In the experiments, we have stored the data for each pond for 4 hours 10 minutes. Among the 250 minutes' data, we have considered data after 3 minutes as we only consider the data for the stable circuit.

## 3.5. DO, COD and BOD

In addition, we find out the biochemical oxygen demand, chemical oxygen demand, and dissolved oxygen of water in pond 1.

Dissolved Oxygen (DO) is oxygen that is dissolved in water. After experiment the DO, we get the 6.79 mg/L shown in Figure 3.10.

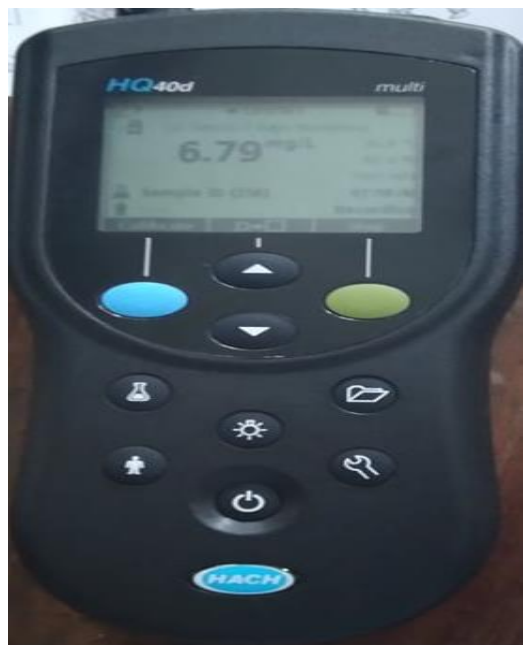

Figure 3.10: DO measurement

Chemical Oxygen Demand (COD) is the amount of oxygen necessary to oxidize all soluble and insoluble organic compounds existing in a volume of water. Its value is usually stated in milligrams per liter of water (mg/L). For calculating the value of COD, in the vial of COD, The sample (S) and distilled (D) are dissolved in 2 ml of the COD vial and kept in the COD reactor at 150 degrees centigrade for 2 hours. The vial is kept in a wooden box for 5-6 hours to cool. To get the COD value, I will make **D** to 0 in the spectrophotometer machine and take the **S** as a reading. Thus we get the COD values. We get the 12 milligrams per liter of water (mg/L) as COD value.

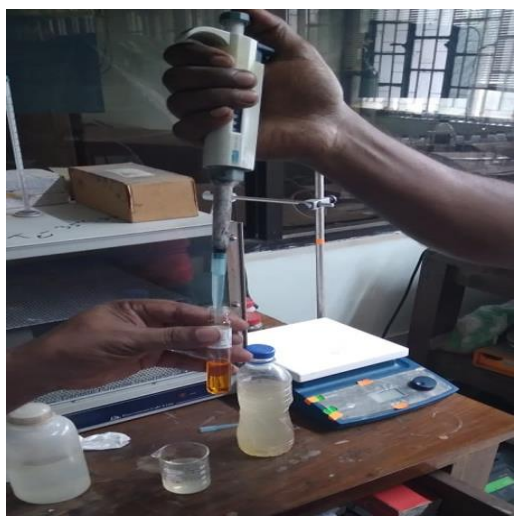
(a) Mixing D in vial

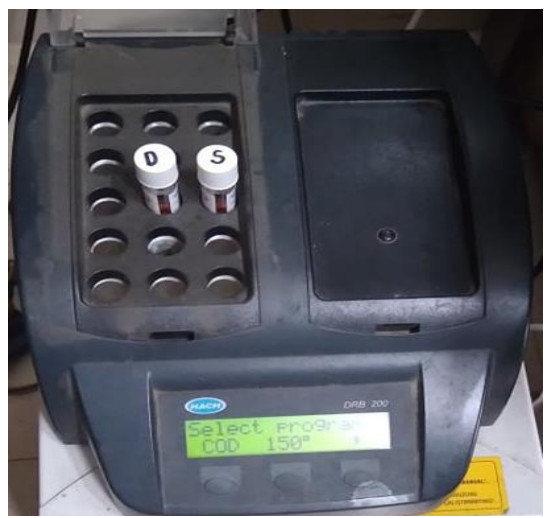
(b) COD Reactor

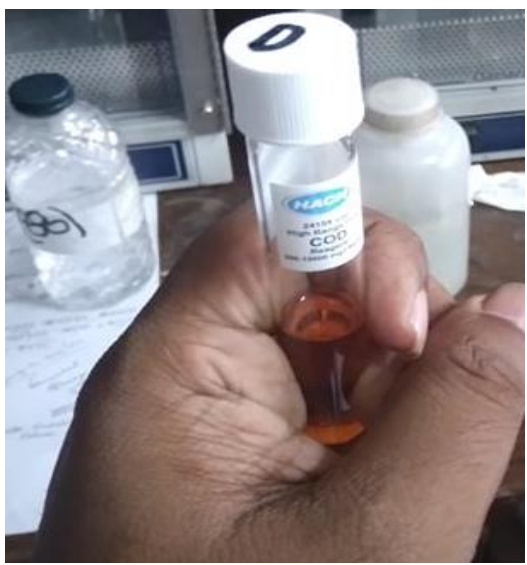
(c) Distilled

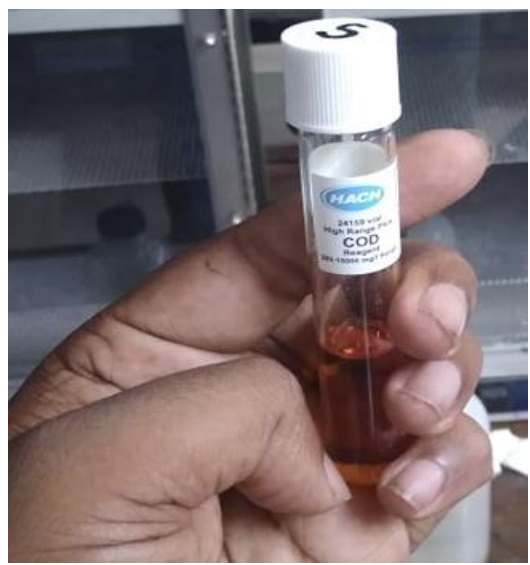
(d) Sample

Figure 3.11: COD instruments

For microbial breakdown (oxidation), Biochemical Oxygen Demand (BOD) quantifies the quantity of oxygen required or consumed in water. For measuring the BOD values, we take the sample according to the COD values. Because BOD depends on COD values. In the OxiTop bottle, we keep the sample of BOD and there is a black dropper in the cap of the bottle. Then, two granules of sodium hydroxide are placed in it. After then, we do the zero level of bottle scale. After labeling zero the OxiTop, it is kept in the reactor of BOD at the constant 20 degrees centigrade for 5 days. This is $BOD_5$ test. Thus, we get the BOD values. After experiment the BOD, we get the 7 mg/L.

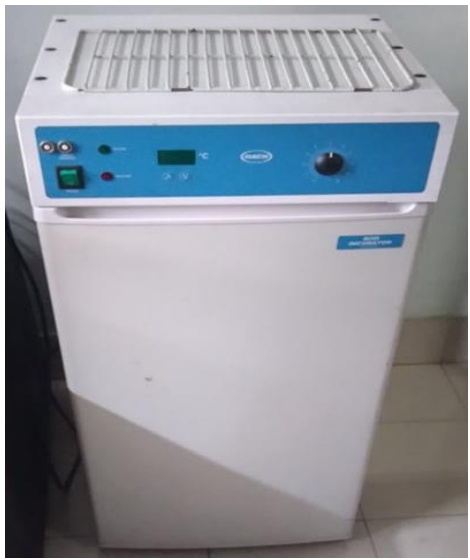
(a) BOD Reactor

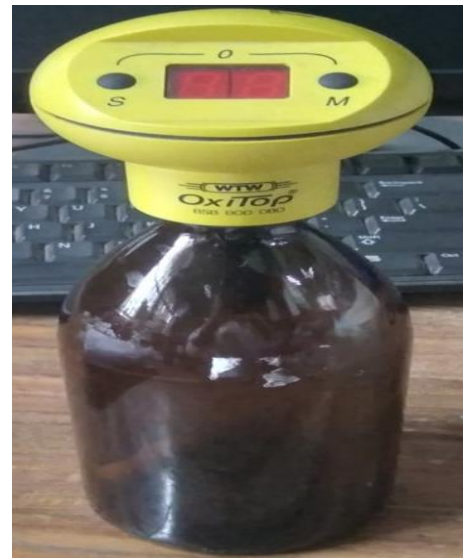
(b) OxiTop Bottle

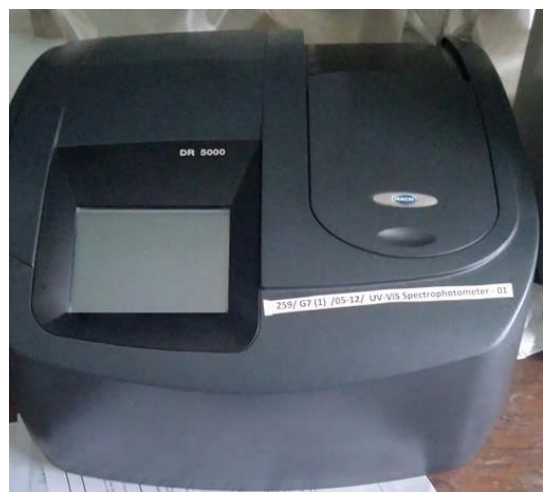
(c) Spectrophotometer machine

Figure 3.12: BOD instruments

In this chapter, we can summarize that all experimental setup for getting the real-time values as well as the proposed methodology is demonstrated here. In addition, we also explained all the water quality parameters sensors along with the cloud server network. Also, in one scenario, we measured BOD, COD and DO. The following chapter will describe the data validation using 10 machine learning models.

# Chapter 4

# Machine Learning Model Analysis

In this chapter, we analyzed the real-time dataset using machine learning models. We used different types of machine learning algorithms for analyzing the gotten real-time dataset. We used J48, Random Forest, K-Nearest Neighbors, K* Algorithms, PART, Decision Table, JRIP, LMT, Logit Boost, REPTree Algorithms. We use Weka software for analyzing algorithms.

## 4.1. Overview

Figure 4.1 describes the detailed block diagram of machine learning classifiers. We prepare our real data with fish categories. Aquatic parameters are the independent variable and fish categories are the dependent variable.

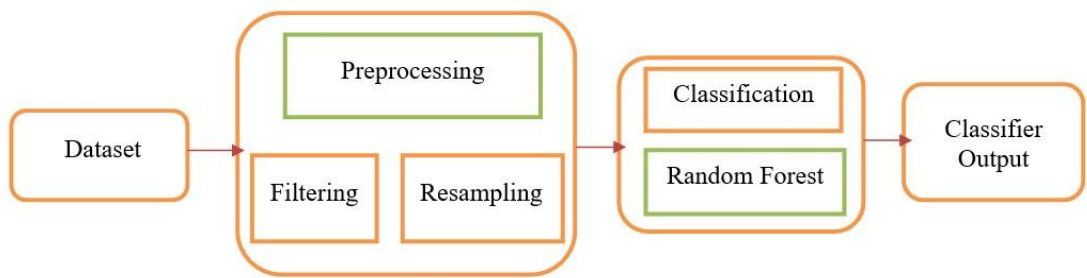

Figure 4.1: Block diagram of machine learning classifiers

## 4.2. Performance Metrics For Classification

We used the WEKA tool to analyze the data set in this section. The Waikato Environment for Knowledge Analysis (Weka) is an ML algorithms software suite established by New Zealand's University of Waikato. It is engraved in the Java programming language. We spoke about success metrics for review in this section. In this step, we get the final result performance of all executed machine learning models. By coding machine learning models, we can exam the results. In this part, we can observe the detailed accuracy of running all stated models. We measured accuracy, precision, recall, and f1-score. Moreover, we also measured the weighted and macro average of the scores for each classifier. All of these terms are obtained from the confusion matrix which is an N × N matrix used for assessing the presence of a cataloging model, where N is the number of target classes. The matrix associates the real goal values with the machine model. This offers us a picture of how and what kind of mistakes our classification model performs. A normal sight of the confusion matrix is shown in Table 4.1.

TABLE 4.1  Confusion Matrix

|  | Predicted Yes | Predicted No |
|---|---|---|
| Actual Yes | True Positive | False Negative |
| Actual No | False Positive | True Negative |

- True Positive: The number of positive events or cases of a data set that are correctly predicted is denoted as TP.

- True Negative: The number of negative events or cases of a data set that are correctly predicted is denoted as TN.

- False Negative: FP is the measure of positive cases that are predicted incorrectly as negative.

- False Positive: FN is the measure of negative cases that are predicted incorrectly as positive.

- Accuracy: Accuracy is the measure of correctly classified cases of a data set. It is expressed mathematically in equation 4.1.

$$\text{Accuracy} = \frac{\text{True Positive} + \text{True Negative}}{All\ cases} \quad (4.1)$$

- Precision (P): P is the proportion of positive cases accurately anticipated for

the positive totals. The low false-positive rate is highly accurate. It is a measure of the exactness of a classifier. It is defined mathematically in equation 4.2.

$$P = \frac{TP}{TP + FP} \qquad (4.2)$$

- Recall (R): R is the ratio of correctly predicted positive cases to all predicted positive cases of a classifier. It is a measure of the completeness of a classifier. R is defined mathematically in equation 4.3.

$$R = \frac{TP}{TP + FN} \qquad (4.3)$$

- F1-Score: The F1-score is calculated as the weighted average of Precision and Recall. In most cases, F1 is more useful than accuracy, when there is uneven class distribution in the data set. It is shown mathematically in equation 4.4.

$$F1 - score = \frac{2 \times (P \times R)}{P + R} \qquad (4.4)$$

- Weighted Average (WA): It is the weighted average of the sum of the scores (Precision / Recall/ F1- score) of all classes after multiplying their respective class proportion. For example, the equation of WA for precision is shown in equation 4.5.

$$WA_P = \frac{\sum_{i=0}^{N-1} P_{Class\_i} \times D_{Class\_i}}{D} \qquad (4.5)$$

## 4.3. Analysis of Machine Learning Models

We know, there are types of machine learning models.

- Supervised Learning
- Unsupervised Learning
- Reinforcement Learning

### 4.3.1. Supervised Learning

Supervised learning is teaching like a human child. In supervised learning, the machine is taught by example with a labeled dataset. Two types of problems can be solved with supervised learning.

- Regression: Numerical value is predicted through regression. For example, if we want to know how many degrees Celsius the average temperature can be if the humidity of the air is 7 tomorrow.
- Classification: Different classes are predicted through classification. For example, some of the emails that come to our emails come as spam. The computer can recognize a new mail, whether the mail is normal mail or spam. This is how spam filtering or classification is done.

### 4.3.2. Unsupervised Learning

In the case of unsupervised learning, no predictions are made. In this case, there is no need to teach the machine with a labeled dataset. Unsupervised learning identifies different hidden similar groups inside machine data. For example, a computer from a supermarket customer data will divide customers into different groups based on different features. Maybe those who shop more are in one group and those who shop less are in another group. Maybe those who buy more food and beverages are in one group and those who buy more luxury items are in one group. Through unsupervised learning, the machine thus discovers the various groups within the data.

### 4.3.3. Reinforcement Learning

Reinforcement learning is mainly applied in self-drive cars, robots, various AI games, etc. In this method, the machine learns on its own through the goal-reward method. In this case, a goal is set for the machine, if the machine can achieve it in the right way, it gets rewarded, in this way the machine keeps learning, if it works in any way, it will be able to fulfill the goal quickly. And thus the efficiency of the machine also increases with time. In the thesis, we used supervised classification models.

## 4.4. Analysis Using Classification Algorithms

We prepared our dataset [Appendix B] after collecting real-time values using sensors from the cloud server of our system. In addition, we labeled the dataset according to the fish categories including katla, sing, prawn, rui, koi, pangas, tilapia, silvercarp, carpio, magur, etc. The optimum temperature of growing fish is (20-26)°c, (15-25)°c, (18-30)°c, and (20-25)°c respectively for rui fish, koi fish, silvercarp fish, and karpio fish [64-67]. During the analysis, we used fish as a target variable and water parameters as feature values. The result of data analysis using ten machine learning models is described below step by step.

### 4.4.1. J48 Classification Algorithm

The algorithm starts by building a decision tree out of the available training data's attribute values. It recognizes the attribute that specifically distinguishes the different instances when it happenstances a collection of items -training set. This function provides information about data instances so that we can identify them appropriately. It is claimed to provide the most information benefit. If there is a number for this function for which no uncertainty is available, i.e. all data instances in the category have the same value for the target variable, this value is ended and the target value we acquired is assigned to it [68]. Figure 4.2 shows the resulting screenshot of the J48 model taken from Weka software.

```
Classifier output

Time taken to build model: 0.01 seconds

=== Stratified cross-validation ===
=== Summary ===

Correctly Classified Instances         533               90.1861 %
Incorrectly Classified Instances        58                9.8139 %
Kappa statistic                          0.8871
Mean absolute error                      0.0205
Root mean squared error                  0.1221
Relative absolute error                 12.9674 %
Root relative squared error             43.4131 %
Total Number of Instances              591

=== Detailed Accuracy By Class ===

               TP Rate  FP Rate  Precision  Recall  F-Measure  MCC    ROC Area  PRC Area  Class
               0.983    0.004    0.966      0.983   0.974      0.972  0.990     0.966     katla
               0.796    0.030    0.709      0.796   0.750      0.727  0.946     0.762     sing
               0.500    0.009    0.583      0.500   0.538      0.530  0.955     0.550     prawn
               0.960    0.012    0.941      0.960   0.950      0.940  0.988     0.968     rui
               0.667    0.003    0.833      0.667   0.741      0.740  0.895     0.616     koi
               0.859    0.021    0.859      0.859   0.859      0.838  0.959     0.887     pangas
               0.930    0.019    0.930      0.930   0.930      0.911  0.978     0.944     tilapia
               0.927    0.004    0.962      0.927   0.944      0.939  0.970     0.921     silverCup
               0.818    0.004    0.931      0.818   0.871      0.866  0.936     0.866     karpio
               1.000    0.003    0.846      1.000   0.917      0.918  0.998     0.866     magur
               0.980    0.002    0.980      0.980   0.980      0.978  0.989     0.962     shrimp
Weighted Avg.  0.902    0.013    0.903      0.902   0.901      0.889  0.971     0.903
```

Figure 4.2: Result screenshot of J48 model taken from Weka software

Figure 4.2 presents average TP Rate as 0.902, FP Rate as 0.013, Precision as 0.903, Recall as 0.902, F-measure as 0.901, MCC as 0.889, ROC Area as 0.971, PRC Area as 0.903, Correctly Classified Instances as 90.19%, Incorrectly Classified Instances as 9.81%, Kappa statistics as 89%, mean absolute error as 2.05%, Root mean squared (RMS) error as 12.21%, Relative absolute error (RAE) as 12.97%, Root relative squared error as 43.41%.

### 4.4.2. Random Forest

Random forest is a kind of democratic algorithm. In this algorithm, the decision is made by voting. Such an algorithm is called ensemble learning. Random forests are made up of many trees or trees. Just as there are many trees in the forest, there are many decision trees in the random forest. The decision that most trees make is considered the final decision [69]. Figure 4.3 shows the resulting screenshot of the Random Forest model taken from Weka software.

```
Classifier output

Time taken to build model: 0.08 seconds

=== Stratified cross-validation ===
=== Summary ===

Correctly Classified Instances         558               94.4162 %
Incorrectly Classified Instances        33                5.5838 %
Kappa statistic                          0.9359
Mean absolute error                      0.02
Root mean squared error                  0.0914
Relative absolute error                 12.6234 %
Root relative squared error             32.5067 %
Total Number of Instances              591

=== Detailed Accuracy By Class ===

               TP Rate  FP Rate  Precision  Recall  F-Measure  MCC    ROC Area  PRC Area  Class
               0.966    0.006    0.949      0.966   0.957      0.953  0.989     0.979     katla
               0.918    0.015    0.849      0.918   0.882      0.872  0.990     0.938     sing
               0.929    0.012    0.650      0.929   0.765      0.771  0.990     0.794     prawn
               0.980    0.002    0.990      0.980   0.985      0.982  0.994     0.989     rui
               0.733    0.002    0.917      0.733   0.815      0.816  0.992     0.881     koi
               0.923    0.008    0.947      0.923   0.935      0.925  0.989     0.973     pangas
               0.953    0.006    0.976      0.953   0.965      0.955  0.996     0.989     tilapia
               0.945    0.002    0.981      0.945   0.963      0.959  0.988     0.965     silverCup
               0.909    0.009    0.857      0.909   0.882      0.876  0.997     0.967     karpio
               0.909    0.000    1.000      0.909   0.952      0.953  1.000     0.992     magur
               0.980    0.000    1.000      0.980   0.990      0.989  1.000     0.999     shrimp
Weighted Avg.  0.944    0.006    0.949      0.944   0.945      0.939  0.993     0.972
```

Figure 4.3: Result screenshot of Random Forest model taken from Weka software

Average TP Rate as 0.944, FP Rate as 0.006, Precision as 0.949, Recall as 0.944, F-measure as 0.945, MCC as 0.939, ROC Area as 0.993, PRC Area as 0.972, Correctly Classified Instances as 94.4161%, Incorrectly Classified Instances as 5.5838%, Kappa statistics as 93.59%, mean absolute error as 2%, RMS error as 9.14%, RAE as 12.62%, Root relative squared error as 32.50% are seen in Figure 4.3.

### 4.4.3. K-Nearest Neighbors (K-NN)

K-NN is a non-parametric distance-based algorithm. The algorithm computes the K number of distances between neighboring data points and discovers the best K for the dataset. It is used for regression as well as classification, but it is mostly used for classification issues [70]. KNN is regarded as an uninterested learner calculation because it does not gain from the preparation set immediately, but rather supplies the dataset and performs an activity on it at the time of order.

```
Classifier output

Time taken to build model: 0 seconds

=== Stratified cross-validation ===
=== Summary ===

Correctly Classified Instances         552               93.401 %
Incorrectly Classified Instances        39                6.599 %
Kappa statistic                          0.9243
Mean absolute error                      0.0135
Root mean squared error                  0.1088
Relative absolute error                  8.523 %
Root relative squared error             38.6865 %
Total Number of Instances              591

=== Detailed Accuracy By Class ===

               TP Rate  FP Rate  Precision  Recall  F-Measure  MCC    ROC Area  PRC Area  Class
               0.897    0.011    0.897      0.897   0.897      0.885  0.965     0.884     katla
               0.918    0.017    0.833      0.918   0.874      0.863  0.969     0.832     sing
               0.929    0.002    0.929      0.929   0.929      0.927  0.990     0.884     prawn
               0.970    0.004    0.980      0.970   0.975      0.970  0.985     0.969     rui
               0.800    0.007    0.750      0.800   0.774      0.769  0.970     0.698     koi
               0.923    0.010    0.935      0.923   0.929      0.918  0.979     0.935     pangas
               0.938    0.006    0.976      0.938   0.957      0.945  0.977     0.960     tilapia
               0.964    0.004    0.964      0.964   0.964      0.960  0.987     0.945     silverCup
               0.848    0.005    0.903      0.848   0.875      0.868  0.966     0.830     karpio
               0.909    0.007    0.714      0.909   0.800      0.802  0.984     0.603     magur
               1.000    0.000    1.000      1.000   1.000      1.000  1.000     1.000     shrimp
Weighted Avg.  0.934    0.007    0.937      0.934   0.935      0.927  0.979     0.920
```

Figure 4.4: Result screenshot of K-NN model taken from Weka software

From Figure 4.4, K-NN model shows average TP Rate as 0.934, FP Rate as 0.007, Precision as 0.938, Recall as 0.934, F-measure as 0.935, MCC as 0.927, ROC Area as 0.979, PRC Area as 0.920, Correctly Classified Instances as 93.40%, Incorrectly Classified Instances as 5.599%, Kappa statistics as 0.92, mean absolute error as 1.35%, RMS error as 10.88%, RAE as 8.523%, Root relative squared error as 38.6865%.

### 4.4.4. K* Algorithm

K* is a classifier based on an instance, expressed as the resemblance of the class of training instances[71].

```
Classifier output

Time taken to build model: 0 seconds

=== Stratified cross-validation ===
=== Summary ===

Correctly Classified Instances         531               89.8477 %
Incorrectly Classified Instances        60               10.1523 %
Kappa statistic                          0.8837
Mean absolute error                      0.041
Root mean squared error                  0.1252
Relative absolute error                 25.8822 %
Root relative squared error             44.5057 %
Total Number of Instances              591

=== Detailed Accuracy By Class ===

              TP Rate  FP Rate  Precision  Recall  F-Measure  MCC    ROC Area  PRC Area  Class
              0.948    0.013    0.887      0.948   0.917      0.908  0.997     0.979     katla
              0.878    0.020    0.796      0.878   0.835      0.820  0.988     0.899     sing
              0.786    0.005    0.786      0.786   0.786      0.781  0.996     0.849     prawn
              0.980    0.014    0.933      0.980   0.956      0.947  0.998     0.994     rui
              0.733    0.005    0.786      0.733   0.759      0.753  0.986     0.797     koi
              0.910    0.037    0.789      0.910   0.845      0.823  0.989     0.943     pangas
              0.829    0.002    0.991      0.829   0.903      0.884  0.988     0.973     tilapia
              0.945    0.002    0.981      0.945   0.963      0.959  0.984     0.961     silverCup
              0.848    0.007    0.875      0.848   0.862      0.854  0.995     0.950     karpio
              0.818    0.003    0.818      0.818   0.818      0.815  0.982     0.849     magur
              0.940    0.004    0.959      0.940   0.949      0.945  1.000     0.995     shrimp
Weighted Avg. 0.898    0.012    0.905      0.898   0.899      0.887  0.992     0.957
```

Figure 4.5: Result screenshot of K* model taken from Weka software

Figure 4.5 presents average TP Rate as 0.898%, FP Rate as 0.012%, Precision as 0.905, Recall as 0.898, F-measure as 0.899, MCC as 0.887, ROC Area as 0.992, PRC Area as 0.957, Correctly Classified Instances as 89.85%, Incorrectly Classified Instances as 10.15%, Kappa statistics as 88.37%, mean absolute error as 4.1%, RMS error as 12.52%, RAE as 25.88%, Root relative squared error as 44.50%.

### 4.4.5. Logistic Model Tree (LMT) Algorithm

LMT is the classification tree that includes logistic regression functions on the leaves, the classifier is used. The method supports target variables in binary and multi-classes, number, name, and absent value[72].

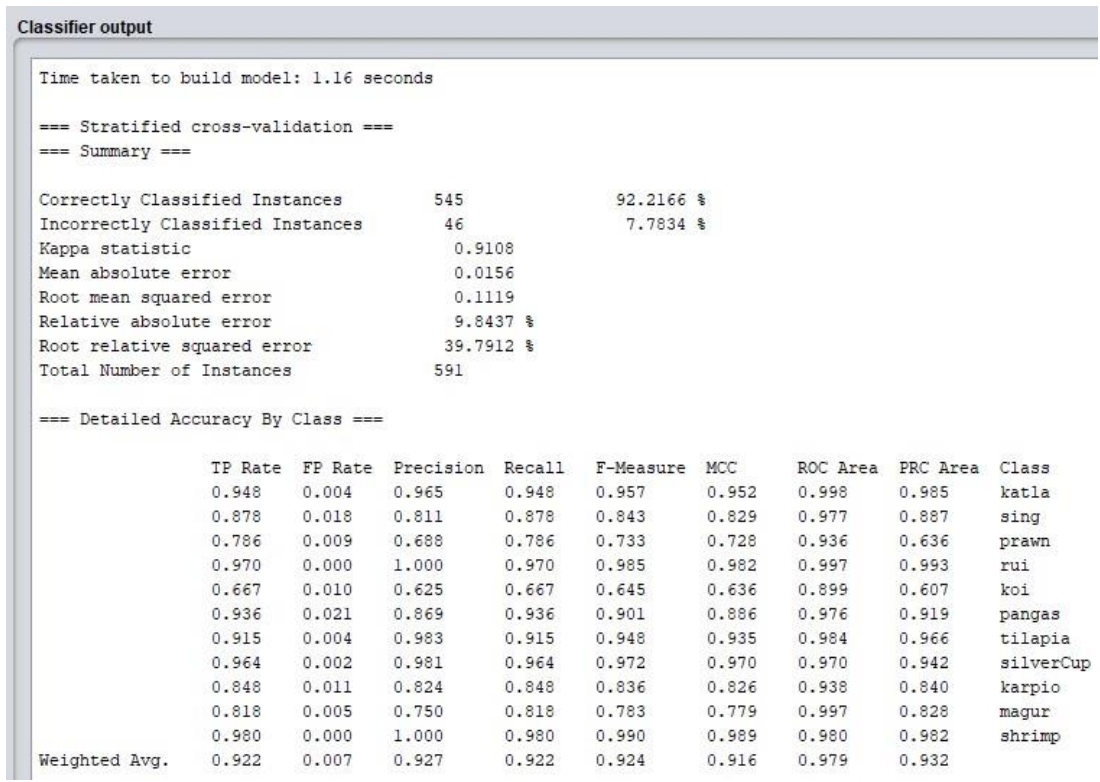

Figure 4.6: Result screenshot of LMT model taken from Weka software

Figure 4.6 presents average TP Rate as 0.922, FP Rate as 0.007, Precision as 0.927, Recall as 0.922, F-measure as 0.924, MCC as 0.916, ROC Area as 0.979, PRC Area as 0.932, Correctly Classified Instances as 88.48%, Incorrectly Classified Instances as 11.52%, Kappa statistics as 91.08%, mean absolute error as 1.56%, RMS error as 11.19%, RAE as 9.84%, Root relative squared error as 39.79%.

### 4.4.6. Reduced-Error Pruning Tree (REPTree)

REPTree is yet another Weka-specific algorithm. It is a quick decision tree learner which has been optimized for simplicity and speed. Reduced-error pruning with backfitting is used in the algorithms to find the smallest representation of the most accurate subtree for the pruning set [73].

```
Classifier output

Time taken to build model: 0.01 seconds

=== Stratified cross-validation ===
=== Summary ===

Correctly Classified Instances         496               83.9255 %
Incorrectly Classified Instances        95               16.0745 %
Kappa statistic                          0.8155
Mean absolute error                      0.0353
Root mean squared error                  0.1463
Relative absolute error                 22.2948 %
Root relative squared error             52.0336 %
Total Number of Instances              591

=== Detailed Accuracy By Class ===

              TP Rate  FP Rate  Precision  Recall  F-Measure  MCC     ROC Area  PRC Area  Class
              0.931    0.019    0.844      0.931   0.885      0.873   0.996     0.966     katla
              0.592    0.030    0.644      0.592   0.617      0.585   0.963     0.760     sing
              0.286    0.016    0.308      0.286   0.296      0.280   0.940     0.315     prawn
              0.980    0.022    0.898      0.980   0.937      0.925   0.992     0.980     rui
              0.667    0.016    0.526      0.667   0.588      0.580   0.888     0.612     koi
              0.744    0.018    0.866      0.744   0.800      0.775   0.948     0.822     pangas
              0.891    0.015    0.943      0.891   0.916      0.894   0.979     0.957     tilapia
              0.891    0.017    0.845      0.891   0.867      0.854   0.976     0.917     silverCup
              0.758    0.014    0.758      0.758   0.758      0.743   0.924     0.714     karpio
              0.636    0.010    0.538      0.636   0.583      0.577   0.949     0.598     magur
              0.960    0.002    0.980      0.960   0.970      0.967   0.998     0.985     shrimp
Weighted Avg. 0.839    0.017    0.841      0.839   0.838      0.822   0.972     0.882
```

Figure 4.7: Result screenshot of REPTree model taken from Weka software

Figure 4.7 presents average TP Rate as 0.839, FP Rate as 0.017, Precision as 0.841, Recall as 0.839, F-measure as 0.838, MCC as 0.822, ROC Area as 0.972, PRC Area as 0.882, Correctly Classified Instances as 83.93%, Incorrectly Classified Instances as 16.07%, Kappa statistics as 81.55%, mean absolute error as 3.53%, RMS error as 14.63%, RAE as 22.29%, Root relative squared error as 52.03%.

### 4.4.7. JRIP

JRip (RIPPER) is a basic and widely used algorithm. Classes are examined as to their size increases, and An initial set of class rules is constructed using incremental error reduction. JRip (RIPPER) begins by giving all examples of a specific decision in the training data as a class and determining protocols that covers all members class. It then moves on to the next class and repeats the process until all classes have been enclosed [74].

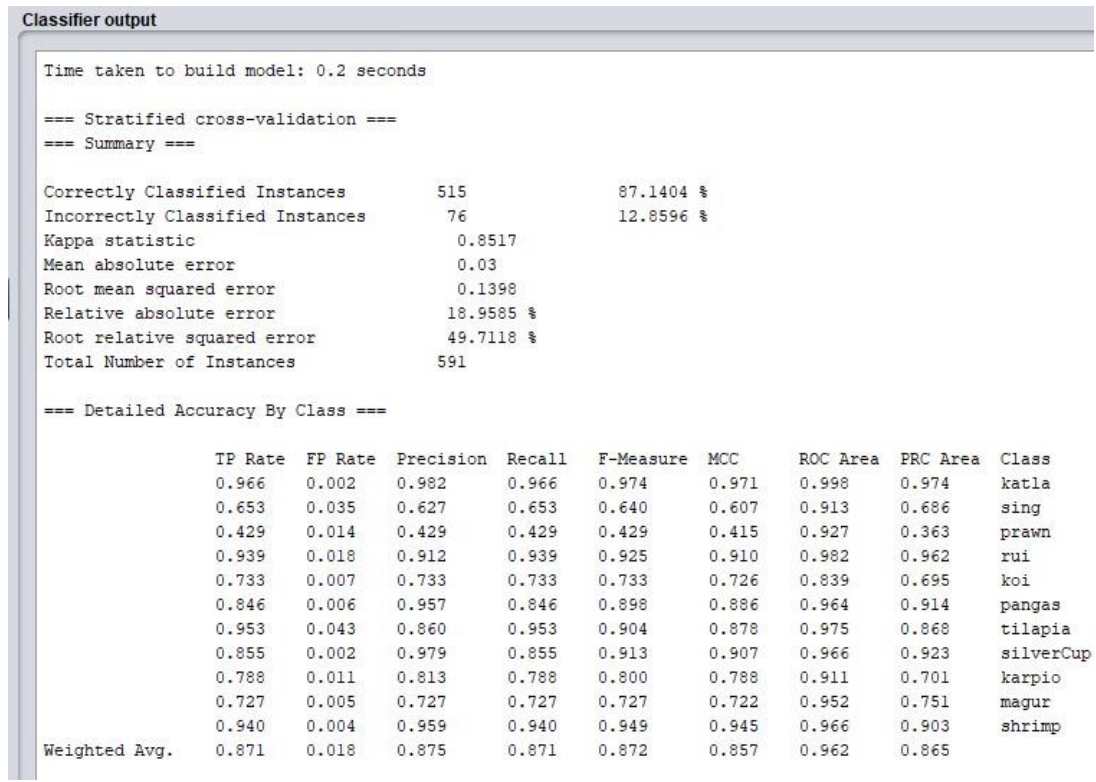

Figure 4.8: Result screenshot of JRIP model taken from Weka software

Figure 4.8 presents average TP Rate as 0.871, FP Rate as 0.018, Precision as 0.875, Recall as 0.871, F-measure as 0.872, MCC as 0.857, ROC Area as 0.962, PRC Area as 0.865, Correctly Classified Instances as 87.14%, Incorrectly Classified Instances as 12.86%, Kappa statistics as 85.17%, mean absolute error as 3%, RMS error as 13.98%, RAE as 18.96%, Root relative squared error as 49.71%.

### 4.4.8. PART

PART is a law learner who divides and conquers. It produces "decision lists," which are defined rules. The new data is compared to each rule in the list in turn and a class of the first matching rule is applied to the object. In each iteration, PART constructs a partial C4.5 decision tree and converts the "best" leaf into a rule [75].

```
Classifier output

Time taken to build model: 0.03 seconds

=== Stratified cross-validation ===
=== Summary ===

Correctly Classified Instances         534               90.3553 %
Incorrectly Classified Instances        57                9.6447 %
Kappa statistic                          0.8892
Mean absolute error                      0.0189
Root mean squared error                  0.1223
Relative absolute error                 11.9376 %
Root relative squared error             43.4993 %
Total Number of Instances              591

=== Detailed Accuracy By Class ===

                 TP Rate  FP Rate  Precision  Recall  F-Measure  MCC    ROC Area  PRC Area  Class
                 0.931    0.009    0.915      0.931   0.923      0.915  0.964     0.917     katla
                 0.816    0.024    0.755      0.816   0.784      0.765  0.949     0.778     sing
                 0.643    0.009    0.643      0.643   0.643      0.634  0.957     0.584     prawn
                 0.980    0.012    0.942      0.980   0.960      0.953  0.988     0.962     rui
                 0.667    0.005    0.769      0.667   0.714      0.709  0.862     0.574     koi
                 0.872    0.016    0.895      0.872   0.883      0.866  0.947     0.870     pangas
                 0.922    0.015    0.944      0.922   0.933      0.915  0.978     0.937     tilapia
                 0.909    0.006    0.943      0.909   0.926      0.919  0.979     0.938     silverCup
                 0.788    0.004    0.929      0.788   0.852      0.848  0.921     0.805     karpio
                 1.000    0.007    0.733      1.000   0.846      0.853  0.997     0.752     magur
                 1.000    0.002    0.980      1.000   0.990      0.989  0.999     0.980     shrimp
Weighted Avg.    0.904    0.012    0.905      0.904   0.903      0.892  0.967     0.893
```

Figure 4.9: Result screenshot of PART model taken from Weka software

Figure 4.9 presents average TP Rate as 0.904, FP Rate as 0.012, Precision as 0.905, Recall as 0.904, F-measure as 0.903, MCC as 0.892, ROC Area as 0.967, PRC Area as 0.893, Correctly Classified Instances as 90.35%, Incorrectly Classified Instances as 9.64%, Kappa statistics as 88.92%, mean absolute error as 1.89%, RMS error as 12.23%, RAE as 11.94%, Root relative squared error as 43.50%.

### 4.4.9. Decision Table

Each class should have its own set of decision rules. A decision table is commonly used to describe the rules. Rough sets can be used to pick attribute subsets as well. Weka classifiers are used to find the algorithm decision table in Rules [76]. The best way to describe machine learning output is to view it in the same format as the input.

```
Classifier output

Time taken to build model: 0.07 seconds

=== Stratified cross-validation ===
=== Summary ===

Correctly Classified Instances         476               80.5415 %
Incorrectly Classified Instances       115               19.4585 %
Kappa statistic                          0.7755
Mean absolute error                      0.1157
Root mean squared error                  0.2148
Relative absolute error                 73.0676 %
Root relative squared error             76.3939 %
Total Number of Instances              591

=== Detailed Accuracy By Class ===

                 TP Rate  FP Rate  Precision  Recall   F-Measure  MCC      ROC Area  PRC Area  Class
                 0.948    0.060    0.632      0.948    0.759      0.746    0.982     0.844     katla
                 0.551    0.011    0.818      0.551    0.659      0.648    0.936     0.578     sing
                 0.429    0.005    0.667      0.429    0.522      0.526    0.964     0.442     prawn
                 0.949    0.045    0.810      0.949    0.874      0.851    0.969     0.875     rui
                 0.467    0.005    0.700      0.467    0.560      0.563    0.891     0.423     koi
                 0.782    0.019    0.859      0.782    0.819      0.794    0.949     0.848     pangas
                 0.822    0.050    0.822      0.822    0.822      0.772    0.964     0.905     tilapia
                 0.909    0.011    0.893      0.909    0.901      0.891    0.979     0.894     silverCup
                 0.727    0.002    0.960      0.727    0.828      0.828    0.937     0.810     karpio
                 0.818    0.014    0.529      0.818    0.643      0.650    0.905     0.559     magur
                 0.740    0.002    0.974      0.740    0.841      0.837    0.990     0.901     shrimp
Weighted Avg.    0.805    0.030    0.821      0.805    0.803      0.782    0.961     0.823
```

Figure 4.10: Result screenshot of decision table model taken from Weka software

Figure 4.10 presents average TP Rate as 0.805, FP Rate as 0.030, Precision as 0.821, Recall as 0.805, F-measure as 0.803, MCC as 0.782, ROC Area as 0.961, PRC Area as 0.823, Correctly Classified Instances as 80.54%, Incorrectly Classified Instances as 19.46%, Kappa statistics as 77.55%, mean absolute error as 11.57%, RMS as 21.48%, RAE as 73.07%, Root relative squared error as 76.39%.

### 4.4.10. Logitboost Algorithm

This class is intended for the regression of additives. This class is classified by regression as a basic learner and can tackle difficulties in several classes[77].

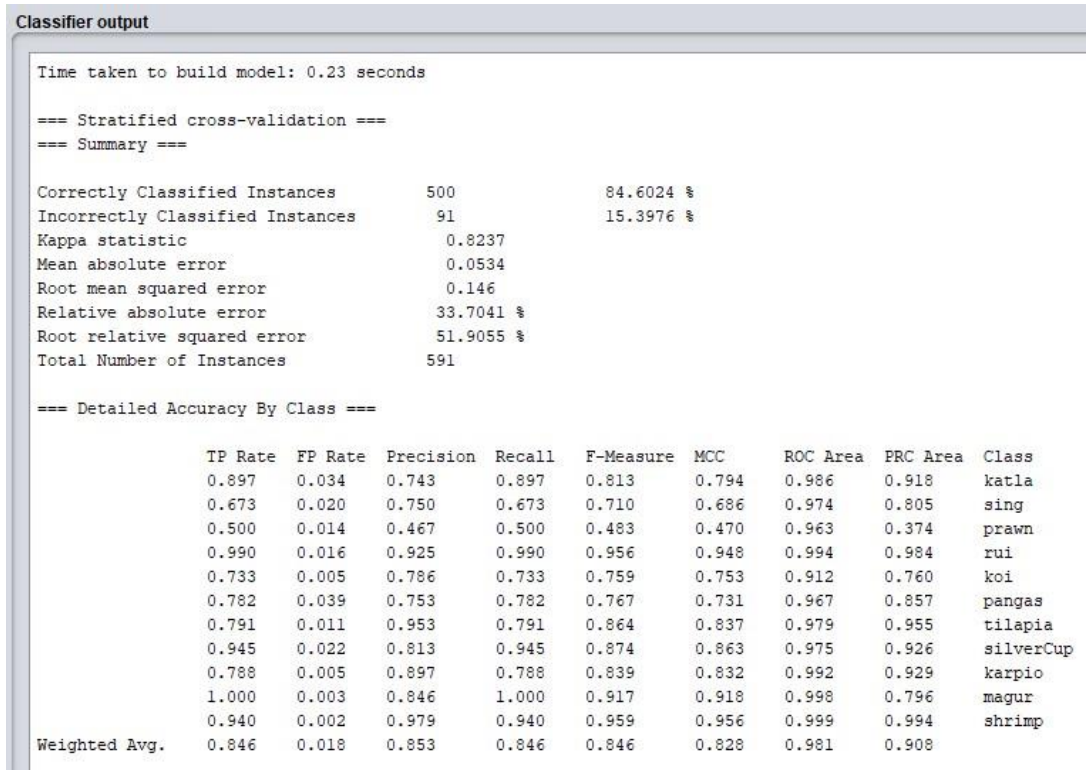

Figure 4.11: Result screenshot of Logitboot model taken from Weka software

Figure 4.11 presents average TP Rate as 0.846, FP Rate as 0.018, Precision as 0.853, Recall as 0.846, F-measure as 0.846, MCC as 0.828, ROC Area as 0.981, PRC Area as 0.908, Correctly Classified Instances as 84.60%, Incorrectly Classified Instances as 15.40%, Kappa statistics as 82.37%, mean absolute error as 5.34%, RMS error as 14.6%, RAE as 33.70%, Root relative squared error as 51.90%.

From the above discussion, it can be summarised that this chapter focuses on the real-time values which are received from our system using machine learning algorithms. All schemas are systematically demonstrated for the performance metrics including accuracy, kappa statistics, TP rate, precision, and so on. The following chapter will describe the discussion and result of the analysis.

# Chapter 5

## Discussion and Result Analysis

In this chapter, result in analysis and discussion on different parameters of aquatic environment like pH, temperature, turbidity, and conductivity have been focused. All necessary graphs for all the parameters of the aquatic environment are deliberated here. Ponds selection and comparison among the performance metrics using machine learning algorithms are also demonstrated here.

## 5.1. Result Analysis of real-time values from the experimental setup

In this section, we have presented all graphs of the value of each parameter of all ponds. 20 samples of real values are taken to represent horizontally and standard reference values of each parameter are presented vertically. Different colors are marked for different ponds for understanding easily.

### 5.1.1. Result Analysis of real-time values of pH

Figure 5.1 shows the graph of the pH value of the real-time data. The range of pH value is 6.02-8.39, 8.57-8.87, 6.00-7.83, 6.51-8.30 and 3.84-3.95 for pond-1, pond-2, pond-3, pond-4, and pond-5, respectively. Figure 5.5 shows the pH values of all ponds from the real environment. The pH range of pond-1 is suited for fish production according to the standard ideal value (6.5-8.5). The received pH value, 8.57-8.87 from pond-2 is greater than the ideal value. Therefore, this pond is not perfect for fish farming. From figure 3.10, pond-3 exhibited the range of pH is 6.00-7.83. It is almost near to ideal range, 6.5-8.5. Pond-4 also provides acceptable values of pH. The range of pH values is not perfect for fish farming for pond-5. This causes to die for fish species. It is a death point for acidity.

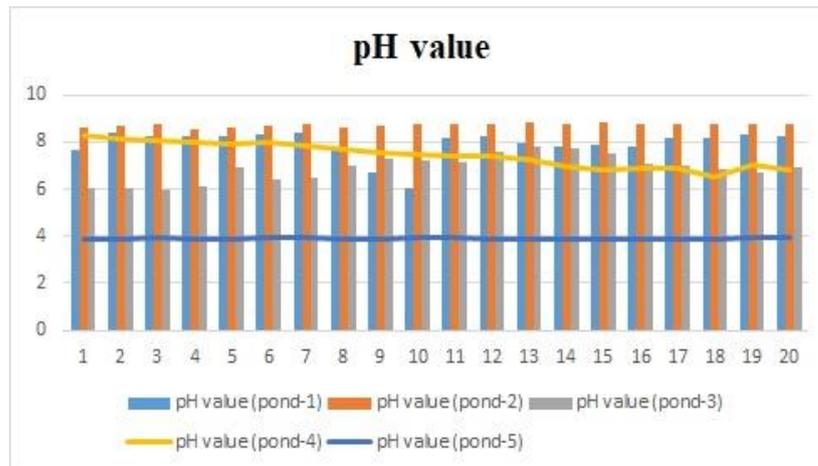

Figure 5.1: Real-time value of pH of all experiments

### 5.1.2. Result Analysis of real-time values of Turbidity

Figure 5.2 displays a graph of real-time values of turbidity from all of the experiments. The range of turbidity values is 3.55-3.57 NTU, 3.41-3.50 NTU, 3.31-3.49 NTU, 3.60-3.62 NTU, and 3.56-3.58 NTU for pond-1, pond-2, pond-3, pond-4, and pond-5, respectively. The acceptable turbidity range is below 10 NTU for fish farming according to the ideal value. Figure 5.2 dictates that all received values of turbidity from all experiments are acceptable.

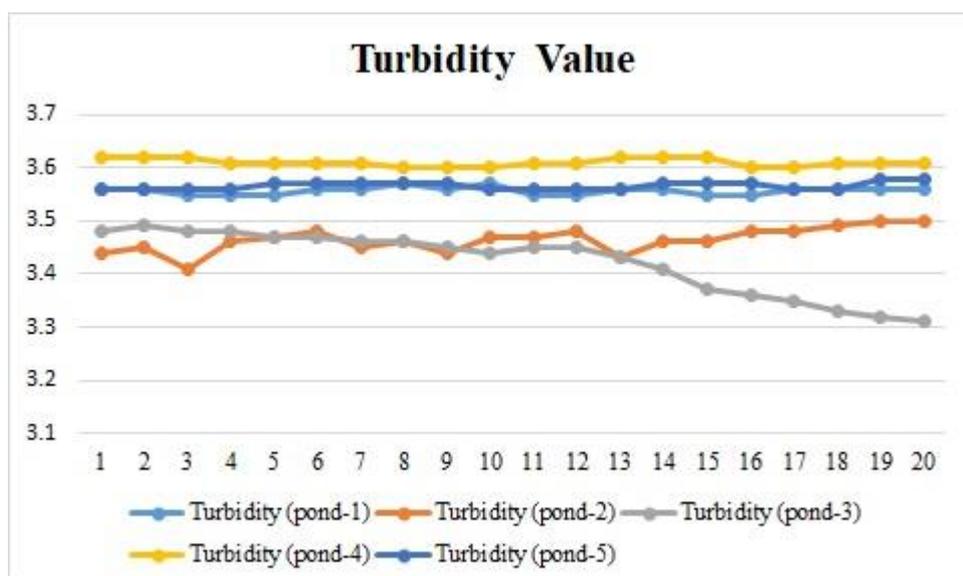

Figure 5.2: Real-time value of turbidity of all experiments

### 5.1.3. Result Analysis of real-time values of Temperature

Figure 5.3 displays a graph of real-time values of temperature from all of the experiments. The range of temperature values is 17.50-17.75 °c, 17.75-18.00 °c, 20.87-21.06 °c, 21.06-21.44 °c, and 21.06-21.25 °c for pond-1, pond-2, pond-3, pond-4, and pond-5, respectively. Temperature varies for various fish species. The overall acceptable range of temperature for a pond is (16-24)°c. The received real-time sensor of all experiments is perfect for fish farming.

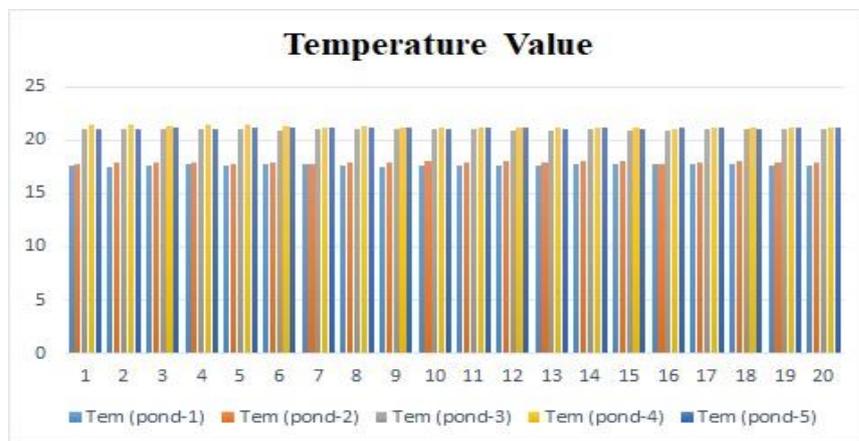

Figure 5.3: Real-time value of temperature of all experiments

### 5.1.4. Result Analysis of real-time values of Conductivity

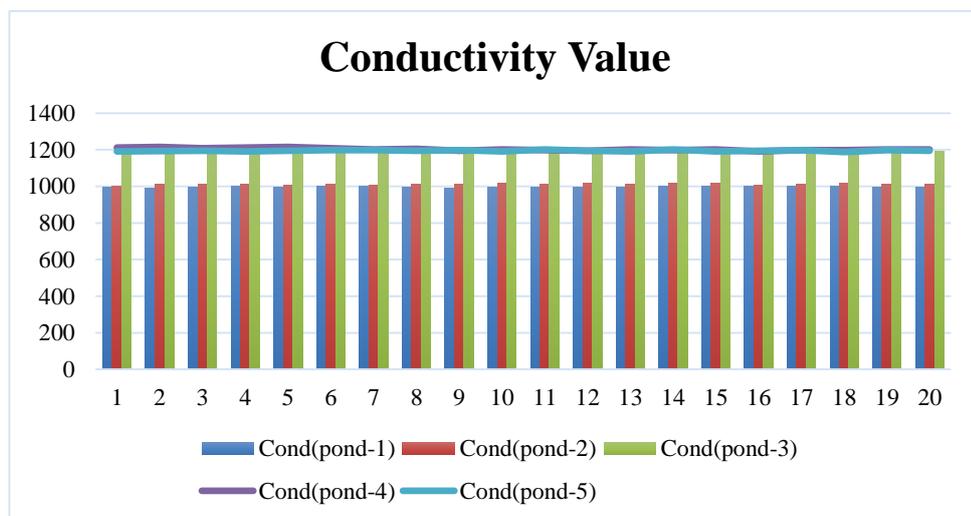

Figure 5.4: Real-time value of conductivity of all experiments

Conductivity is a helpful parameter to assess the virtue of water. It is subject to ionic fixation and water temperature. Conductivity is additionally viewed as a sign of its newness or in any case of a water body [78]. The temperature in a water body is directly connected to its conductivity. At 25 Degrees-C, has a specific conductance of 1,413 mmhos/cm [79]. When water temperature increases, conductivity increases. For every 1°C increase, conductivity values can increase 2-4%. Temperature affects conductivity by increasing ionic mobility [80]. The ideal range of it is 970-1825 μS/cm.

Figure 5.4 displays a graph of real-time values of conductivity from all of the experiments. The range of conductivity values is 989-1003 μS/cm, 1003-1017 μS/cm, 1179-1190 μS/cm, 1193-1215 μS/cm, and 1190-1201 μS/cm for pond-1, pond-2, pond-3, pond-4, and pond-5, respectively. The received real-time sensor of all experiments is perfect for fish farming.

## 5.2. Pond Suggestion

Here we have shown all the real-time values of aquatic environmental quality parameters from all ponds.

Table 5.1 illustrates the summary of all received real-time values for each pond. The ideal range of all parameters including pH: 6.5-8.5, turbidity: below 10 ntu, temperature: (16-24)°c, and the depth of the pond should not be less than 1 meter or more than 5 meters, the best depth is 2 meters for fish farming in the pond shown in Table 2.1.

TABLE 5.1  Received values for suggesting pond

| Pond | pH | Temperature ($^o$C) | Turbidity (ntu) | Depth (m) | Conductivity (μS/cm) | Remarks |
|---|---|---|---|---|---|---|
| Pond 1 | 6.02-8.39 | 17.50-17.75 | 3.55-3.57 | 1-2 | 989-1003 | Recommended |
| Pond 2 | 8.57-8.87 | 17.75-18.00 | 3.41-3.50 | 1-2 | 1003-1017 | Not Recommended |
| Pond 3 | 6.00-7.83 | 20.87-21.06 | 3.31-3.49 | 1-2 | 1179-1190 | Recommended |
| Pond 4 | 6.51-8.30 | 21.06-21.44 | 3.60-3.62 | 2-4 | 1193-1215 | Recommended |
| Pond 5 | 3.84-3.95 | 21.06-21.25 | 3.56-3.58 | 1-3 | 1190-1201 | Not recommended |

According to this record, pond-1, pond-3, and pond-4 are eligible for fish farming. The pond-2 is not perfect for cultivation fish. Because of the pH value, 8.57-8.87 is higher than the ideal values. It is the point of the slow growth of fish. The pond-5 is not suited for the lower pH value, 3.84-3.95. This causes fish death. That is why we could not give suggestions for pond-1 and pond-2 to fish farmers.

## 5.3. Comparison Performance Metrics

The received real-time values are analyzed using 10 machine learning algorithms. Table 5.2 shows the comparison of performance metrics including accuracy, kappa statistics and avg. TP rate among the classifiers.

TABLE 5.2  Comparison among classification model based on the performance metric

| S.L. No | Machine Learning Model | Accuracy (%) | Kappa Statistics (%) | Avg. TP Rate (%) | Position |
|---|---|---|---|---|---|
| 1 | J48 | 90.19 | 88.7 | 90.2 | 5$^{th}$ Rank |
| 2 | Random Forest | 94.42 | 93.5 | 94.4 | **1$^{st}$ Rank** |
| 3 | K-NN | 93.4 | 92.4 | 93.4 | 2$^{nd}$ Rank |
| 4 | K* Algorithm | 89.85 | 88.37 | 89.8 | 6$^{th}$ Rank |
| 5 | LMT | 92.22 | 91.08 | 92.2 | 3$^{rd}$ Rank |
| 6 | REPTree | 83.93 | 81.5 | 83.9 | 9$^{th}$ Rank |
| 7 | JRIP | 87.14 | 85.17 | 87.1 | 7$^{th}$ Rank |
| 8 | PART | 90.35 | 88.92 | 90.4 | 4$^{th}$ Rank |
| 9 | Decision Table | 80.54 | 77.5 | 80.5 | **10$^{th}$ Rank** |
| 10 | Logit boost | 84.60 | 82.37 | 84.6 | 8$^{th}$ Rank |

Table 5.2 shows, Random forest gives the highest score of every metric as accuracy 94.42%, kappa statistic as 93.5.11%, and Avg. True Positive (TP) rate as 94.4%. The second highest score belongs to the KNN model which tells accuracy as 93.4%, kappa statistic as 92.4%, and TP rate as 93.4%. LMT acquires the 3rd highest position by achieving an accuracy as 92.22%, kappa statistic as 91.08% and TP rate as 92.2%. PART has 4th place in scoring performance metrics by getting accuracy of 90.35%, kappa statistic as 88.92 and TP rate as 90.4%. J48 takes 5$^{th}$ place as getting accuracy as 90.195, kappa statistics as 88.7%, and Avg. TP Rate as 90.2%. K* gives the 6$^{th}$ score of every metric as accuracy 89.85%, kappa statistic as 88.37%, and Avg. True Positive (TP) rate as 89.8%. The 7$^{th}$ highest score belongs to the JRIP model which tells accuracy as 87.14%, kappa statistic as 85.17%, and Avg. TP rate as 87.1%. Logit Boost acquires the 8$^{th}$ highest position by achieving an accuracy as 84.60%, kappa statistic as 82.37% and TP rate as 84.6%. REPTree has 9$^{th}$ place in scoring performance metrics by getting accuracy as 83.93%, kappa statistic as 81.5% and TP rate as 83.9%. The decision table gives the lowest score as getting accuracy of 80.54%, kappa statistics as 77.5%, and Avg. TP Rate as 80.5%.

Random forest gives the highest score for all performance metrics. Because we know that it is a kind of democratic algorithm. In this algorithm, the decision is made by voting. Such an algorithm is called ensemble learning. There are many decision trees in the random forest, the decision that most trees make is considered the final decision.

The graphical representation of this table is shown in Figure 5.5. Accuracy is indicated by a blue-colored line, kappa statistics is marked by a dark red-colored line. And the average true positive (TP) rate is demonstrated by the olive-colored line.

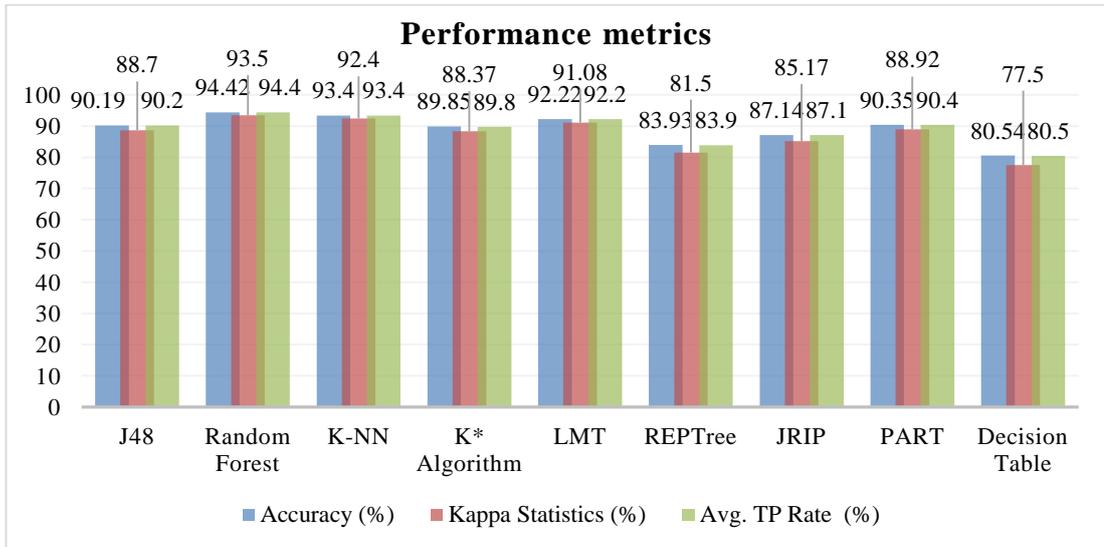

Figure 5.5: Graphical presentation of the comparison

This chapter illustrates the different results and discussion. This chapter also depicts the pond scenarios using several aquatic parameters as well as the comparison of the prediction using the machine learning models. Among five ponds, 3 ponds are satisfied for cultivating the fish, and the rest 2 pond is not satisfied with the standard reference values.

# Chapter 6

# Conclusion and Future Works

## 6.1. Conclusion and Future Work

An IoT framework for real-time data monitoring is proposed and positively instigated using diverse sensors, and circuits as validated in different figures. We experimented using 5 scenarios of ponds for obtaining real-time values including pH, temperature, conductivity, and turbidity values. After analyzing the real-time values, we found that pond-1, pond-3, and pond-4 are perfect in fish farming where the water quality of these ponds satisfied by achieving pH, Temperature, Turbidity, Conductivity, and Depth. And pond-2 and pond-5 are not perfect for fish farming. That is why, end-user can cultivate fish in pond-1, pond-3, and pond-4. For pond-2 and pond-5, a farmer can take any step for utilizing these ponds for fish farming later. As the proposed model is computer well-ordered, farmers will be able to properly monitor their ponds. The implementation enables the sensor to provide data to the thingSpeak server. The data were analyzed using ten ML algorithms. Among the executed ML algorithms, Random Forest took first place with accuracy 94.42%, kappa statistics 93.5%, and Avg. TP rate 94.4%. We also investigated the BOD, COD and DO for one scenario.

The Limitation of the thesis is as follows:
- Only 5 ponds water quality were measured.
- Free cloud network was used
- Only four sensors were used; BOD and DO sensors are very expensive [81-82]
- Machine learning algorithms were applied

In the future, the proposed architecture may be tested with more aquatic environment parameters by targeting specific fish species. We will experiment in a river or lake-canal. We can buy a premium cloud server for storing data. Then, we will apply deep learning, transfer learning, or any updated technologies.

# Reference

[1] FAO, The State of World Fisheries and Aquaculture 2020. [Online] from http://www.fao.org/3/ca9231en/CA9231EN.pdf.

[2] C.H.S. Ruxton, et al., "The health benefits of omega-3 polyunsaturated fatty acids: a review of the evidence," *Journal of Human Nutrition and Dietetics*, vol. 17, pp. 449-459, https://doi.org/10.1111/j.1365-277X.2004.00552.x, 2004.

[3] https://www.who.int/news-room/fact-sheets/detail/the-top-10-causes-of-death. Accessed 5 November 2020.

[4] L. Djoussé, et al., "Fish consumption, omega-3 fatty acids and risk of heart failure: a meta-analysis," *Clin Nutr.,* vol. 31, no. 6, pp.846-53, doi: 10.1016/j.clnu.2012.05.010, 2012.

[5] J. Zheng, et al., "Fish consumption and CHD mortality: an updated meta-analysis of seventeen cohort studies," *Public Health Nutr,* vol. 15, no. 4, pp.725-37, doi: 10.1017/S1368980011002254, 2012.

[6] R. Chowdhury, et al., "Association between fish consumption, long chain omega 3 fatty acids, and risk of cerebrovascular disease: systematic review and meta-analysis," *BMJ*, doi: 10.1136/bmj.e6698. PMID: 23112118; PMCID: PMC3484317.

[7] S. Buscemi, et al., "Habitual fish intake and clinically silent carotid atherosclerosis," *Nutr J,* vol. 13, no. 2, doi: 10.1186/1475-2891-13-2. PMID: 24405571; PMCID: PMC3893519.

[8] J. C. McCann and B. N. Ames BN, "Is docosahexaenoic acid, an n-3

long-chain polyunsaturated fatty acid, required for development of normal brain function? An overview of evidence from cognitive and behavioral tests in humans and animals," *Am J Clin Nutr.*, vol. 82, no. 2, pp. 281-95, doi: 10.1093/ajcn.82.2.281. PMID: 16087970, 2005.

[9]  B. Koletzko, et al., "The roles of long-chain polyunsaturated fatty acids in pregnancy, lactation and infancy: review of current knowledge and consensus recommendations," *J Perinat Med,* vol. 36, no. 1, pp.5-14, doi: 10.1515/JPM.2008.001. PMID: 18184094, 2008.

[10] M.C. Morris MC, et al., "Fish consumption and cognitive decline with age in a large community study," *Arch Neurol,* vol. 62, no. 12, pp.1849-53. doi: 10.1001/archneur.62.12.noc50161, 2005.

[11] https://internetofthingsagenda.techtarget.com/definition/IoT-device. Accessed, 4 November 2020.

[12] https://www.itu.int/en/ITU-T/gsi/iot/Pages/default.aspx. Accessed, 4 November 2020.

[13] Bhatnagar, et al., "Water quality guidelines for the management of pond fish culture," *International journal of environmental sciences*, vol. 3, no.6, pp. 1980-2009, 2013.

[14] https://www.chemeurope.com/en/whitepapers/126405/bod-cod-toc-and-tod-sum-parameters-in-environmental-analysis.html. Accessed: 29 November 2020.

[15] M. Sato, et al., "Measurements of aquatic pollutant loading amounts using the automatic composite sampler and automatic sampler." *Bulletin of Aichi Environmental Research Center (Japan) (1979).*

[16] http://fisheries.tamu.edu/files/2013/09/Dissolved-Oxygen-for-Fish-Production1.pdf, Accessed: 19 June 2021.

[17] https://www.merusonline.com/bod-biological-oxygen-demand/, Accessed: 19 June 2021.

[18] I. Lee, and L. Kyoochun, "The Internet of Things (IoT): Applications, investments, and challenges for enterprises," Business Horizons, vol. 58, no. 4, pp. 431-440, 2015.

[19] Idoje, et al., "Survey for smart farming technologies: Challenges and issues," *Computers & Electrical Engineering*, vol. 92, pp. 107104, 2021.

[20] L. Hang, et al., "A secure fish farm platform based on blockchain for agriculture data integrity," *Computers and Electronics in Agriculture*, vol. 170, pp. 105251, 2020.

[21] https://www.scientechworld.com/it-educational-platforms/iot-solutions/iot-builder, Accessed: 30 May 2021.

[22] https://www.libelium.com/libeliumworld/smart-water-sensors-to-monitor-water-quality-in-rivers-lakes-and-the-sea/. Accessed 5 November 2020

[23] T. Kageyama, et al., "A wireless sensor network platform for water quality monitoring," *IEEE SENSORS*, pp. 1-3, doi: 10.1109/ICSENS.2016.7808887, 2016.

[24] F. Adamo, et al., "A Smart Sensor Network for Sea Water Quality Monitoring," *IEEE Sensors Journal*, vol. 15, no. 5, pp. 2514-2522, doi:

10.1109/JSEN.2014.2360816, 2015.

[25] G. Wiranto, et al., "Integrated online water quality monitoring," *International Conference on Smart Sensors and Application (ICSSA),* pp. 111-115, doi: 10.1109/ICSSA.2015.7322521, 2015.

[26] C. Encinas, et al., "Design and implementation of a distributed IoT system for the monitoring of water quality in aquaculture," *Wireless Telecommunications Symposium (WTS),* pp. 1-7, doi: 10.1109/WTS.2017.7943540, 2017.

[27] M. M. Islam, et al., "Design and Implementation of an IoT System for Predicting Aqua Fisheries Using Arduino and KNN," Lecture Notes in Computer Science, vol 12616. Springer, Cham. https://doi.org/10.1007/978-3-030-68452-5_11. 2021.

[28] M.M. Islam, et al., "Aqua fishing monitoring system using IoT devices," *Int. J. Innov. Sci. Eng. Technol*, vol. 6, no. 11, pp. 109-114, 2019.

[29] L. V. Q. Danh, et al., "Design and Deployment of an IoT-Based Water Quality Monitoring System for Aquaculture in Mekong Delta," *International Journal of Mechanical Engineering and Robotics Research*, vol. 9, no. 8, 2020.

[30] L. Parra, et al., "Design and Deployment of Low-Cost Sensors for Monitoring the Water Quality and Fish Behavior in Aquaculture Tanks during the Feeding Process," *Sensors (Basel),* vol. 18, no. 3, pp.:750. doi: 10.3390/s18030750. PMID: 29494560; PMCID: PMC5877200, 2018.

[31] G.I. Hapsari, et al., "IoT-based guppy fish farming monitoring and

controlling system," Telkomnika, vol. 18, no. 3, pp. 1538-1545, 2020.

[32]　T. Alam, et al., "A Data-Driven Deployment Approach for Persistent Monitoring in Aquatic Environments," *Second IEEE International Conference on Robotic Computing (IRC)*, pp. 147-154, doi: 10.1109/IRC.2018.00030, 2018.

[33]　M. Ahmed, et al., "Analyzing the Quality of Water and Predicting the Suitability for Fish Farming based on IoT in the Context of Bangladesh," *International Conference on Sustainable Technologies for Industry 4.0 (STI)*, pp. 1-5, doi: 10.1109/STI47673.2019.9068050, 2019.

[34]　L. Hang, et al., "A secure fish farm platform based on blockchain for agriculture data integrity," *Computers and Electronics in Agriculture*, vol. 170, pp. 105251, 2020.

[35]　P. Sun and Y. Chen, "Aquiculture Remote Monitoring System Based on Internet of Things," *International Conference on Robots & Intelligent System (ICRIS)*, pp. 187-190, doi: 10.1109/ICRIS.2019.00056, 2019.

[36]　D. M. M. Khan, "An IoT Based Smart Water Monitoring System for Fish Firming in Bangladesh," doi:10.3390/ECWS-5-08044, 2020.

[37]　J. Lee, et al., "Realization of Water Process Control for Smart Fish Farm," *International Conference on Electronics, Information, and Communication (ICEIC)*, pp. 1-5, doi: 10.1109/ICEIC49074.2020.9051285, 2020.

[38]　N. A. J. Salih, et al., "Design and implementation of a smart monitoring system for water quality of fish farms*," Indonesian Journal of Electrical Engineering and Computer Science*, vol.14, no. 1, 44-50.

[39]  A.Ramya, et al., "IOT Based Smart Monitoring System for Fish Farming" *International Journal of Engineering and Advanced Technology (IJEAT)* ISSN: 2249 – 8958, Volume-8 Issue-6S, August 2019.

[40]  W. Sung, et al., "Remote fish aquaculture monitoring system based on wireless transmission technology," *International Conference on Information Science, Electronics and Electrical Engineering*, pp. 540-544, 2014.

[41]  D. Prangchumpol, "A Model of Mobile Application for Automatic Fish Feeder Aquariums System," *International Journal of Modeling and Optimization*, vol. 8. Pp. 277-280, 10.7763/IJMO.2018.V8.665, 2018.

[42]  I. Ullah, et al., "An Optimization Scheme for Water Pump Control in Smart Fish Farm with Efficient Energy Consumption," *Processes*, vol. 6, no. 65, 2018.

[43]  G. Bourke, et al., "A decision support system for aquaculture research and management," Aquacultural Engineering, vol. 12, no. 2, pp. 111-123, 1993.

[44]  W. Li, et al., "Prediction of dissolved oxygen in a fishery pond based on gated recurrent unit (GRU)," *Information Processing in Agriculture,* 2020.

[45]  G. A. Defe and A. Z. C. Antonio, "Multi-parameter Water Quality Monitoring Device for Grouper Aquaculture," *IEEE 10th International Conference on Humanoid, Nanotechnology, Information Technology, Communication and Control, Environment and Management (HNICEM)*,

pp. 1-5, doi: 10.1109/HNICEM.2018.8666414, 2018.

[46] T. Abinaya, et al., "A Novel Methodology for Monitoring and Controlling of Water Quality in Aquaculture using Internet of Things (IoT)," *International Conference on Computer Communication and Informatics (ICCCI)*, pp. 1-4, doi: 10.1109/ICCCI.2019.8821988, 2019.

[47] P. Yang, et al., "Prediction of water quality evaluation for fish ponds of aquaculture," *56th Annual Conference of the Society of Instrument and Control Engineers of Japan (SICE),* pp. 545-546, doi: 10.23919/SICE.2017.8105455, 2017.

[48] S. Ogunlana, et al., "Fish classification using support vector machine," *Afr. J. Comput. ICT*, vol. 8, no. 2, pp. 1–8, 2006.

[49] M. M. Fouad, et al., "Automatic Nile Tilapia fish classification approach using machine learning techniques," *13th International Conference on Hybrid Intelligent Systems (HIS 2013),* pp. 173–178, 2013.

[50] W. Silvert, "Decision support systems for aquaculture licensing," *Journal of Applied Ichthyology,* vol. 10, pp. 307–311, 1994.

[51] C. D. Este, et al., "Predicting Shellfish Farm Closures with Class Balancing Methods," *AAI 2012: Advances in Artificial Intelligence, Lecture Notes in Computer Science*, pp. 39–48, 2012.

[52] M. S. Shahriar and A. Rahman, "Spatial-temporal prediction of algal bloom," Ninth International Conference on Natural Computation, pp. 973-977, 2013.

[53] "pH in Fish Farming | Yokogawa Electric

Corporation," www.yokogawa.com. https://www.yokogawa.com/library/resources/application-notes/ph-in-fish-farming/.

[54]    "Pond Water Temperature Guide (Best Pond Thermometers) - Pond Informer," Pond Informer, Mar. 21, 2019. https://pondinformer.com/pond-water-temperature-guide/.

[55]    Fondriest Environmental, Inc. "Turbidity, Total Suspended Solids and Water Clarity." Fundamentals of Environmental Measurements. 13 Jun. 2014. Web. < https://www.fondriest.com/environmental-measurements/parameters/water-quality/turbidity-total-suspended-solids-water-clarity/ >.

[56]    Halim, A., et al. "Assessment of water quality parameters in baor environment, Bangladesh: A review." International Journal of Fisheries and Aquatic Studies 6.2 (2018): 269-263.

[57]    [online]. https://flyby-bd.com/farm-organic-fertilizers-examples/. Accessed: 25 December 2020.

[58]    https://viblo.asia/p/gioi-thieu-ve-arduino-LzD5deOOKjY, Accessed: 30 May 2021.

[59]    http://www.egiman.com/2016/10/esp8266-oc-cam-bien-nhiet-o-esp8266.html, Accessed: 30 May 2021.

[60]    https://how2electronics.com/ph-meter-using-ph-sensor-arduino-oled/, Accessed: 30 May 2021.

[61]    https://i0.wp.com/randomnerdtutorials.com/wp-

content/uploads/2013/11/ultrasonic-sensor-with-arduino-hc-sr04.jpg?quality=100&strip=all&ssl=1, Accessed: 30 May 2021.

[62] https://www.engineersgarage.com/wp-content/uploads/2021/04/Screen-Shot-2021-04-19-at-3.34.13-PM.png, Accessed: 30 May 2021.

[63] https://thingspeak.com/. Accessed 20 December 2020.

[64] R. Kausar and M. Salim, "Effect of Water Temperature on The Growth Performance and Feed Conversion Ratio Of Labeo Rohita," Pakistan Vet. J., vol. 26, no. 3, pp. 105-108, 2006.

[65] https://www.hartz.com/habitat-needs-maintenance-outdoor-koi-pondfish/, accessed: 13 June, 2021.

[66] http://www.fao.org/fishery/docs/DOCUMENT/aquaculture/CulturedSpecies/file/en/en_silvercarp.htm, accessed: 13 June, 2021.

[67] http://www.fao.org/fishery/affris/species-profiles/common-carp/natural-food-and-feeding-habits/en/, accessed: 13 June, 2021.

[68] G. Kaur, et al., "Improved J48 Classification Algorithm for the Prediction of Diabetes," *International Journals of Computer Applications,* vol. 98, no. 22, pp. 13-17. https://doi.org/10.5120/17314-7433, 2014.

[69] M. Belgiu, et al., "Random forest in remote sensing: A review of applications and future directions," *Journal of Photogrammetry and Remote Sensing,* vol. 114, pp. 24-31, https://doi.org/10.1016/j.isprsjprs.2016.01.011.2016.

[70] G. Guo, et al., "KNN model-based approach in classification," *In OTM*

*Confederated International Conferences On the Move to Meaningful Internet Systems,* pp. 986-996, Springer, Berlin, Heidelberg.

[71]  R. Duriqi, et al., "Comparative analysis of classification algorithms on three different datasets using WEKA," *5th Mediterranean Conference on Embedded Computing (MECO),* pp. 335-338, doi: 10.1109/MECO.2016.7525775, 2016.

[72]  M. Sumner, et al., "Speeding up logistic model tree induction," *European conference on principles of data mining and knowledge discovery*, Springer, Berlin, Heidelberg, 2005.

[73]  W. Mohamed, et al., "A comparative study of reduced error pruning method in decision tree algorithms," *IEEE International conference on control system, computing and engineering*, IEEE, 2012.

[74]  R. Anil, et al., "J48 and JRIP rules for e-governance data," *International Journal of Computer Science and Security (IJCSS)*, vol. 5, no. 2, pp. 201, 2011.

[75]  Frank, et al., "Generating accurate rule sets without global optimization," 1998.

[76]  Chen. Cong, et al., "An explanatory analysis of driver injury severity in rear-end crashes using a decision table/Naïve Bayes (DTNB) hybrid classifier," *Accident Analysis & Prevention*, vol. 90, pp. 95-107, 2016.

[77]  Cai. Yu-Dong, et al., "Using LogitBoost classifier to predict protein structural classes," *Journal of theoretical biology*, vol. 238, no. 1, pp. 172-176, 2006.

[78]   K. Dinesh, et al., "Study of seasonal water quality assessment and fish pond conservation in Thanjavur, Tamil Nadu, India," *J. Entomol. Zool. Stud*, vol. 5, pp. 1232-1238, 2017.

[79]   https://www.aquaculturealliance.org/advocate/electrical-conductivity-water-part-1/. Accessed, 29 December 2020.

[80]   https://www.fondriest.com/environmental-measurements/parameters/water-quality/conductivity-salinity-tds/#cond1. Accessed, 29 December 2020.

[81]   https://www.indiamart.com/proddetail/dissolved-oxygen-sensor-2778168233.html?pos=1&pla=n; Accepted: 28 May 2021

[82]   https://www.indiamart.com/proddetail/bod-analysis-apparatus-20522811348.html; Accepted: 28 May 2021

## Appendix A: List of Acronyms

**List of Acronyms**

| | |
|---|---|
| K-NN | K- Nearest Neighbors |
| IoT | Internet of Things |
| REPTree | Reduced Error Pruning Tree |
| LMT | Logistic Model Tree |
| WEKA | Waikato Environment for Knowledge Analysis |
| COD | Chemical Oxygen Demand |

BOD    Biochemical Oxygen Demand

DO    Dissolved Oxygen

# Appendix B: Real-Time Data

## **pH values**

| pH value (pond-1) | pH value (pond-2) | pH value (pond-3) | pH value (pond-4) | pH value (pond-5) |
|---|---|---|---|---|
| 7.67 | 8.65 | 6.02 | 8.3 | 3.84 |
| 8.39 | 8.66 | 6.01 | 8.15 | 3.9 |
| 8.24 | 8.74 | 6 | 8.09 | 3.91 |
| 8.26 | 8.57 | 6.09 | 7.99 | 3.89 |
| 8.23 | 8.62 | 6.93 | 7.92 | 3.88 |
| 8.31 | 8.69 | 6.42 | 7.96 | 3.92 |
| 8.38 | 8.73 | 6.48 | 7.87 | 3.91 |
| 7.84 | 8.63 | 7 | 7.7 | 3.89 |
| 6.7 | 8.66 | 7.27 | 7.54 | 3.9 |
| 6.02 | 8.75 | 7.24 | 7.5 | 3.93 |
| 8.16 | 8.77 | 7.11 | 7.43 | 3.92 |
| 8.23 | 8.78 | 7.6 | 7.42 | 3.86 |
| 7.96 | 8.81 | 7.83 | 7.26 | 3.87 |
| 7.79 | 8.79 | 7.75 | 6.95 | 3.86 |
| 7.88 | 8.87 | 7.5 | 6.78 | 3.9 |
| 7.84 | 8.8 | 7.07 | 6.92 | 3.89 |
| 8.18 | 8.78 | 6.99 | 6.87 | 3.86 |
| 8.16 | 8.8 | 6.87 | 6.51 | 3.85 |
| 8.32 | 8.79 | 6.68 | 7.02 | 3.92 |
| 8.23 | 8.79 | 6.95 | 6.8 | 3.95 |

**Temperature values**

| Temperature (pond-1) | Temperature (pond-2) | Temperature (pond-3) | Temperature (pond-4) | Temperature (pond-5) |
|---|---|---|---|---|
| 17.62 | 17.75 | 21 | 21.44 | 21.06 |
| 17.5 | 17.87 | 21 | 21.5 | 21.09 |
| 17.56 | 17.94 | 21 | 21.37 | 21.12 |
| 17.75 | 17.87 | 21.06 | 21.44 | 21.06 |
| 17.56 | 17.81 | 21 | 21.5 | 21.12 |
| 17.69 | 17.94 | 20.94 | 21.37 | 21.19 |
| 17.69 | 17.81 | 21 | 21.25 | 21.19 |
| 17.56 | 17.94 | 21 | 21.31 | 21.12 |
| 17.5 | 17.94 | 21.06 | 21.12 | 21.19 |
| 17.62 | 18 | 21 | 21.25 | 21.06 |
| 17.62 | 17.94 | 21.06 | 21.19 | 21.25 |
| 17.56 | 18 | 20.94 | 21.12 | 21.12 |
| 17.62 | 17.94 | 20.87 | 21.25 | 21.06 |
| 17.75 | 18 | 21.06 | 21.19 | 21.25 |
| 17.75 | 18 | 20.87 | 21.25 | 21.06 |
| 17.69 | 17.81 | 20.87 | 21.06 | 21.12 |
| 17.69 | 17.87 | 21.06 | 21.19 | 21.19 |
| 17.75 | 18 | 21.06 | 21.19 | 21 |
| 17.56 | 17.94 | 21 | 21.25 | 21.19 |
| 17.56 | 17.94 | 21.06 | 21.25 | 21.12 |

## Conductivity values

| Cond (pond-1) | Cond (pond-2) | Cond (pond-3) | Cond (pond-4) | Cond (pond-5) |
|---|---|---|---|---|
| 995.88 | 1003.23 | 1186.92 | 1211.79 | 1190.31 |
| 989.1 | 1010.01 | 1186.92 | 1215.18 | 1192.01 |
| 992.49 | 1013.97 | 1186.92 | 1207.83 | 1193.7 |
| 1003.23 | 1010.01 | 1190.32 | 1211.79 | 1190.31 |
| 992.49 | 1006.62 | 1186.92 | 1215.18 | 1193.7 |
| 999.83 | 1013.97 | 1183.53 | 1207.83 | 1197.66 |
| 999.83 | 1006.62 | 1186.92 | 1201.05 | 1197.66 |
| 992.49 | 1013.97 | 1186.92 | 1204.44 | 1193.7 |
| 989.1 | 1013.97 | 1190.32 | 1193.7 | 1197.66 |
| 995.88 | 1017.36 | 1186.92 | 1201.05 | 1190.31 |
| 995.62 | 1013.97 | 1190.32 | 1197.66 | 1201.05 |
| 992.49 | 1017.36 | 1183.53 | 1193.7 | 1193.7 |
| 995.88 | 1013.97 | 1179.57 | 1201.05 | 1190.31 |
| 1003.23 | 1017.36 | 1190.32 | 1197.66 | 1201.05 |
| 1003.23 | 1017.36 | 1179.57 | 1201.05 | 1190.31 |
| 999.83 | 1006.62 | 1189.57 | 1190.31 | 1193.7 |
| 999.83 | 1010.01 | 1190.32 | 1197.66 | 1197.66 |
| 1003.23 | 1017.36 | 1190.32 | 1197.66 | 1186.92 |
| 992.49 | 1013.97 | 1186.92 | 1201.05 | 1197.66 |
| 992.49 | 1013.97 | 1190.32 | 1201.05 | 1193.7 |

## **Turbididty values**

| Turbidity (pond-1) | Turbidity (pond-2) | Turbidity (pond-3) | Turbidity (pond-4) | Turbidity (pond-5) |
|---|---|---|---|---|
| 3.56 | 3.44 | 3.48 | 3.62 | 3.56 |
| 3.56 | 3.45 | 3.49 | 3.62 | 3.56 |
| 3.55 | 3.41 | 3.48 | 3.62 | 3.56 |
| 3.55 | 3.46 | 3.48 | 3.61 | 3.56 |
| 3.55 | 3.47 | 3.47 | 3.61 | 3.57 |
| 3.56 | 3.48 | 3.47 | 3.61 | 3.57 |
| 3.56 | 3.45 | 3.46 | 3.61 | 3.57 |
| 3.57 | 3.46 | 3.46 | 3.6 | 3.57 |
| 3.56 | 3.44 | 3.45 | 3.6 | 3.57 |
| 3.57 | 3.47 | 3.44 | 3.6 | 3.56 |
| 3.55 | 3.47 | 3.45 | 3.61 | 3.56 |
| 3.55 | 3.48 | 3.45 | 3.61 | 3.56 |
| 3.56 | 3.43 | 3.43 | 3.62 | 3.56 |
| 3.56 | 3.46 | 3.41 | 3.62 | 3.57 |
| 3.55 | 3.46 | 3.37 | 3.62 | 3.57 |
| 3.55 | 3.48 | 3.36 | 3.6 | 3.57 |
| 3.56 | 3.48 | 3.35 | 3.6 | 3.56 |
| 3.56 | 3.49 | 3.33 | 3.61 | 3.56 |
| 3.56 | 3.5 | 3.32 | 3.61 | 3.58 |
| 3.56 | 3.5 | 3.31 | 3.61 | 3.58 |

# Appendix C: List of Publications

**List of Publications**

1. **Md. Monirul Islam,** Mohammod Abul Kashem, Jia Udiin, "An IoT Framework for Real-time Aquatic Environment Monitoring using an Arduino and Sensors" IAES International Journal of Electrical and Computer Engineering (IJECE), Indonesia, Scopus Indexed, Q2 ranked (Accepted, May 2021)

2. **Md. Monirul Islam,** Mohammod Abul Kashem, Jia Uddin, "Fish Survival Prediction in an Aquatic Environment Using Random Forest Model," IAES International Journal of Artificial Intelligence (IJ-AI), Indonesia, Scopus Indexed, Q2 ranked.

3. **Md. Monirul Islam**, Jia Uddin, Mohammod Abul Kashem, Fazly Rabbi, Md. Waliul Hasnat (2021) Design and Implementation of an IoT System for Predicting Aqua Fisheries Using Arduino and KNN. In: Singh M., Kang DK., Lee JH., Tiwary U.S., Singh D., Chung WY. (eds) Intelligent Human Computer Interaction. IHCI 2020. Lecture Notes in Computer Science, vol 12616. Springer, Cham. https://doi.org/10.1007/978-3-030-68452-5_11

4. Fazly Rabbi, Shihab Uddin Tareq, **Md. Monirul Islam**, Mohammad Asaduzzaman Chowdhury and Mohammod Abul Kashem, "A Multivariate Time Series Approach for Forecasting of Electricity Demand in Bangladesh Using ARIMAX Model," 2020 2nd International Conference on Sustainable Technologies for Industry 4.0 (STI), Dhaka, Bangladesh, 2020, pp. 1-5, doi: 10.1109/STI50764.2020.9350326.

5. Shobnom Mustary, M. Abul Kashem, M. Nurul Islam Khan, Faruk Ahmed Jewel, Md. Monirul Islam and Saiful Islam, "LEACH Based WSN Classification Using Supervised Machine Learning Algorithm," 2021 International Conference on Computer Communication and Informatics (ICCCI), 2021, pp. 1-5, doi: 10.1109/ICCCI50826.2021.9457001.